\documentclass[twocolumn]{article}

\usepackage{abstract}
\usepackage{PRIMEarxiv}
\usepackage{authblk}   % For author/affiliation formatting
\usepackage[utf8]{inputenc} % allow utf-8 input
\usepackage[T1]{fontenc}    % use 8-bit T1 fonts
\usepackage{booktabs}       % professional-quality tables
\usepackage{amsfonts}       % blackboard math symbols
\usepackage{nicefrac}       % compact symbols for 1/2, etc.
\usepackage{microtype}      % microtypography
\usepackage{lipsum}         % garbage text
\usepackage{fancyhdr}       % header
\usepackage{graphicx}       % graphics
\usepackage[numbers]{natbib} % references        
\usepackage[hyphens]{url}   % simple URL typesetting
\usepackage{hyperref}       % hyperlinks
\usepackage{amsmath}        % equations
\usepackage{float}          % to force figures to apper where I want
\usepackage{placeins}
\usepackage{enumitem}       % to kill hanging indents in two columns
\setlist[enumerate,1]{
    label=\textbf{\arabic*)},
    labelsep=0.2em,           % spacing between label and text
    leftmargin=0pt,           % remove left margin
    itemindent=0pt,           % no indent on first line
    labelwidth=0pt,           % don't reserve space for label
    align=left,               % left-align label
    listparindent=0pt,        % no indent for wrapped paragraphs
    parsep=0pt,               % tight paragraph spacing
    itemsep=0.5em             % vertical spacing between items
}

\title{Aesthetics Without Semantics
\thanks{\textbf{\underline{Note}}: 
\textit{Parts of this work were presented in abstract format at the Vision Science of Art Conference (VSAC2016), the Iberian Conference on Perception (CIP2022), and the European Conference on Visual Perception (ECVP2022). See Perception 51, No1 (Suppl.) pp139, 2022)}} 
}

\author[1,2]{C. Alejandro Parraga}
\author[1,2,3]{Olivier Penacchio}
\author[1]{Marcos Muñoz Gonzalez}
\author[2]{Bogdan Raducanu}
\author[1,2]{Xavier Otazu}

\affil[1]{Computer Science Dept., Engineering School, Universitat Autònoma de Barcelona (UAB), Campus UAB, Bellaterra, 08193, Barcelona, Spain}
\affil[2]{Computer Vision Centre, Campus UAB, Bellaterra, 08193, Barcelona, Spain}
\affil[3]{School of Psychology and Neuroscience, University of St Andrews, St Andrews, Fife KY16 9JP, United Kingdom}

\date{}  % No date

\begin{document}
\onecolumn

\maketitle
\begin{abstract}
While it is easy for human observers to judge an image as beautiful or ugly, aesthetic decisions result from a combination of entangled perceptual and cognitive (semantic) factors, making the understanding of aesthetic judgements particularly challenging from a scientific point of view. Furthermore, our research shows a prevailing bias in current databases, which include mostly beautiful images, further complicating the study and prediction of aesthetic responses. We address these limitations by creating a database of images with minimal semantic content and devising, and next exploiting, a method to generate images on the ugly side of aesthetic valuations. The resulting Minimum Semantic Content (MSC) database consists of a large and balanced collection of 10,426 images, each evaluated by 100 observers. We next use established image metrics to demonstrate how augmenting an image set biased towards beautiful images with ugly images can modify, or even invert, an observed relationship between image features and aesthetics valuation. Taken together, our study reveals that works in empirical aesthetics attempting to link image content and aesthetic judgements may magnify, underestimate, or simply miss interesting effects due to a limitation of the range of aesthetic values they consider.
\end{abstract}

% keywords can be removed

\twocolumn

\section{Introduction}
A core challenge in empirical aesthetics is linking subjective aesthetic responses (such as the experience of beauty or emotional arousal), to visual properties of the stimulus and their representations in the brain. In the case of visual aesthetics, a popular “bottom-up” approach assumes that responses are influenced by a combination of low-, mid-, and high-level image features, corresponding to stages in the visual hierarchy. Low-level features include contrast, lightness, saturation, color, etc.~\citep{Carandini2005}; mid-level ones involve grouping, contours, segmentation, depth, etc.~\citep{Freeman2013, Roe2012, Welchman2005} and high-level features relate to object recognition and semantic associations~\citep{Goodale1992}. These studies have shown that features such as color, contrast, saturation, aspect ratio, contour~\citep{Farzanfar2023, Iigaya2021, McManus1980, McManus1981, Reber1998}, image quality, anisotropy, complexity and fractal self-similarity~\citep{Mallon2014, Spehar2003, Tinio2009} are modest predictors of aesthetic preference. Other features that seem to predict preferences are symmetry~\citep{Bertamini2019}, which is preferred on abstract geometrical patterns~\citep{Jacobsen2002} and faces~\citep{Rhodes2006}, and smoothly curved contours, which are preferred over angular contours~\citep{Bar2006, Bertamini2016, Clemente2023, Vartanian2013}.

Despite their success, the explanatory power of feature-based models is limited by small effect sizes, dataset constraints, strong inter-individual variability and methodological variability (for a review see Vessel \textit{et al}~\citep{Vessel2022}). Aesthetic judgments are shaped not only by perceptual input but also by context, memory, emotion and other cognitive factors that vary across individuals and cultures~\citep{Bignardi2021, Martinez2020, Palmer2010, Skov2022, Vessel2010}. This entanglement of perceptual and cognitive factors complicates efforts to predict aesthetic responses from image features alone and is reflected in the strong confounding between shared versus individual visual appeal~\citep{Honekopp2006, Vessel2012, Vessel2013}. In his book “A Brief Tour of Human Consciousness”, best-selling author V.S. Ramachandran states that “assuming that 90\% of the variance seen in art is driven by cultural diversity, this is what people already study and it’s called Art History. As a scientist, I am interested in the 10\% that is universal.”~\citep{Ramachandran2004}. Many researchers that encountered this confounding found that observers do not often agree in what they find visually appealing. In a series of preference study of digital images, Vessel \textit{et al}~\citep{Vessel2018, Vessel2010} found that observers’ judgments revealed a strong degree of shared taste for real-world scenes but were much more idiosyncratic in their ratings of novel, visually diverse images. In other words, the presence of higher-level, unshared semantic meanings and associations led to variability in ratings~\citep{Martinez2020}. A similar result was found for color preferences, which were predicted by participant’s likings of objects associated to those colors~\citep{Palmer2010}.

In a rigorous examination of the sources of variance in aesthetic judgments—whether stemming from the stimuli, the observers, or their interaction—Martinez \textit{et al}.~\citep{Martinez2020} provided estimates that stabilized with repeated measures. Using variance component analysis (VCA) on a limited dataset comprising faces, single objects, and noise patterns, they found that observer-related factors accounted for most of the variability across all cases. However, stimulus characteristics contributed some variance in the case of faces, while stimulus-observer interactions contributed to the cases of both faces and objects. Importantly, they also noted that the results of such analyses depend heavily on the specificity of the hypotheses and the number of stimuli designed to target the relevant variance components.

In light of this situation, we ask to what extent the extensive availability of internet images —often rich in semantic content and laden with inherent biases— has itself contributed to the lack of progress in empirical aesthetics, and what steps might be taken to address this issue. We first show that most existing image aesthetics databases are heavily skewed toward beautiful images and saturated with semantic content, which obscures the true relationships between visual features and aesthetic evaluations. In particular, the beauty bias restricts the dynamic range of ratings and hinders the discovery of genuine correlations between image properties and perceived beauty or ugliness. To address this problem, we introduce a new database specifically designed to balance the aesthetic spectrum and minimize semantic content, enabling a more controlled and reductionist investigation of the perceptual factors underlying aesthetic judgment. Our findings demonstrate that such an approach can reveal or even invert correlations between image features and perceived aesthetics, highlighting the importance of database design in computational aesthetics research.

\section{Challenges in empirical and computational aesthetics}
\label{sec:challenges}

\subsection{Common methodological problems}
The challenge of untangling perceptual and cognitive factors in the putative relationship between image features and aesthetic appreciation emerges already at the level of response measurement and the datasets used for image valuation. Response measurements can be broadly classified into direct and indirect types. Direct measures capture participants’ explicit judgments and preferences, such as binary responses (“yes”/“no”, “ugly”/“beautiful”), Likert scales, continuous rating scales, pairwise comparisons, and free-adjustment sliders~\citep{Belfi2019, Ishizu2011, Jacobsen2006, Kim2007, Vessel2019}. Indirect measures reflect behavioral and physiological correlates of aesthetic engagement, including viewing time, reaction time, arousal levels, pupil dilation, heart rate, and economic proxies like “willingness to pay”, etc.~\citep{Aharon2001, Laeng2016, Lopez2020, Salimpoor2011, Smith2010, Tschacher2012}. A further level of indirectness involves neurophysiological measurements, such as patterns of brain activity associated with different experimental conditions, typically captured via blood oxygen level–dependent (BOLD) signals in fMRI studies~\citep{Bar2007, Boynton1999, Iigaya2023, Jacobsen2006, Keefe2018, Pasupathy2002}. Understandably, all the results were obtained from small collections of images gathered under each author’s criteria (often including art) and typically rated by a small group of participants.
The results of direct measures are difficult to compare—not only because of the interplay between perceptual and cognitive factors, but also due to significant methodological variability. Indirect measures, on the other hand, aim to access unmediated aesthetic appeal—at least partially and without involving conscious reflection—but there is little consensus on which of these measures reliably indicate aesthetic response to visual stimuli. Another unresolved issue is whether incorporating direct measures (e.g., image ratings) during the collection of indirect data (e.g., brain imaging) introduces additional variability across studies. Interpretation is further complicated by the use of different questionnaires to assess aesthetic response—for instance, asking how beautiful, moving, attractive, or likeable an image is—each potentially tapping into distinct yet overlapping constructs. The use of artwork in such research further amplifies cognitive influences, as it is difficult to control for recognition of specific styles (e.g., Picasso, Miró). Such recognition can trigger semantic associations and bias aesthetic judgments, given that prior knowledge of an artwork or artist can strongly shape perceived aesthetic value~\citep{Kirk2009}.

The psychology and neuroscience methods discussed above often rely on simplified or artificial stimuli (e.g., disembodied faces, geometric shapes, patterns, etc.) and small, arbitrarily selected image sets, prioritizing experimental control at the expense of ecological validity. This limitation can be addressed through the use of computer vision techniques and large-scale datasets containing thousands of internet images~\citep{Chounchenani2025, Datta2006, Jiang2010, Joshi2011, Liu2017, Murray2012}. However, identifying consistent correlations between image features and aesthetic judgments in such datasets remains extremely challenging. The same difficulty applies to training AI models to predict aesthetic responses. This reveals an apparent paradox: AI models require large datasets rated by many observers, yet as the datasets become broader and more general, the entanglement between perceptual and cognitive factors in observer responses increases—making it harder to disentangle and model the sources of aesthetic judgment. To address this, computer-based studies often classify the dataset content in a small number of categories (“ugly”/ “beautiful”), create overly-complex models, and sometimes consider an arbitrarily small fraction of the images contained in the datasets, without controlling for the probability of committing a type I error. 

Large image datasets (LIDs) are typically sourced from a limited number of internet platforms, with the most commonly used being Photo.net~\citep{photonet1993}, Flickr~\citep{Flickr2003} and DPChallenge~\citep{DPChallenge2016}. These platforms contain the tools necessary for users to upload digital images and cast votes on different aspects of the images uploaded by other users. Table~\ref{tab:Table01} shows a list of some of the most successful image datasets used in computational aesthetics at the time of writing.

\begin{table*}[htb!]
\centering
\caption{Summary of some of the most popular image datasets used in computational aesthetics studies. All of the databases contain a mixture of natural and human-made objects including semantics-rich representations of people, consumer objects and animals in different contextual situations.}
\label{tab:Table01}
\renewcommand{\arraystretch}{1.2} % add some padding between lines
\begin{tabular}{lllp{6cm}}
\toprule
\textbf{Platform} & \textbf{Database name} & \textbf{Images} & \textbf{Characteristics} \\
\midrule
Photo.net & Photo database~\citep{Datta2006, Joshi2011} & 14,839 & Assessed by an average of 10 observers \\
Photo.net & Photo database~\citep{Datta2010} & 20,000 & Assessed by at least 10 observers \\
Photo.net & Waterloo-IAA~\citep{Liu2017} & 1,100 & Assessed for beauty in a lab-controlled user study (26 observers) \\
DPChallenge.com & DPChallenge~\citep{Datta2006, Joshi2011} & 16,509 & Assessed by an average of 205 observers \\
DPChallenge.com & AVA~\citep{Murray2012} & 250,000 & Assessed by an average of 200 observers \\
DPChallenge.com & CUHK~\citep{Yan2006} & 60,000 & Assessed by at least 100 observers \\
DPChallenge.com & PhotoQuality (CUHK-PQ)~\citep{Luo2011} & 17,690 & Assessed by an average of 200 observers and manually labelled using binary ratings \\
Flickr.com & AROD~\citep{Schwarz2018} & 320,000 & Assessed through time-independent statistics (“faves” and “views”) by thousands of observers \\
Flickr.com & FAE-Captions~\citep{Jin2022} & 251,000 & Assessed through image descriptions \\
Flickr.com & FLICKR-AES~\citep{Ren2017} & 40,000 & Assessed by 210 unique Amazon Mechanical Turk (AMT) annotators \\
Other (Mix) & IAD~\citep{Lu2015} & 1,500,000 & Collected from DPChallenge (300,000 images) and PHOTO.NET (1,200,000 images) \\
Other (Mix) & Kodak Aesthetics~\citep{Jiang2010} & 1,500 & Collected from a number of sources (Flickr, Kodak Picture of the Day, study observers, etc.) and assessed by 30 observers \\
Other & Terragalleria~\citep{Joshi2011} & 14,449 & Contains images of travel photography collected by Dr Quang-Tuan Luong and assessed by an average of 20 observers \\
\bottomrule
\end{tabular}
\end{table*}

Most of the computational studies cited above rely heavily on training artificial neural networks (ANNs) or other classification algorithms to predict aesthetic ratings, with comparatively little attention paid to understanding the underlying image features that drive observers to find an image aesthetically pleasing, emotionally moving, or otherwise engaging. Nevertheless, there are several issues concerning them and the platforms they are based upon:

\begin{itemize}[leftmargin=*, itemsep=0pt, topsep=0pt]
    \item \textbf{Evaluation methods are diverse} and have been implemented using a range of paradigms: namely Absolute Category Rating (observers provide a score based on a fixed set of predefined categories), Likert Scale ratings (observers assess images on a scale of discrete values, usually ranging from 1 to 10) and Automatic Sentiment Analysis of the image comments and descriptions (see for example Jin \textit{et al}.~\citep{Jin2022}; Ren \textit{et al}.~\citep{Ren2017} and Schwarz \textit{et al}.~\citep{Schwarz2018}), which is common in databases derived from Flickr.com. Furthermore, platforms include ratings on several dimensions such as originality, technical merit, aesthetic content, or general appeal and the scores of each category are then averaged to derive an overall aesthetic rating (see for example Datta \textit{et al}.~\citep{Datta2006}). These methodological differences may become an issue in studies that mix images from different sources (see for example Jiang \textit{et al}.~\citep{Jiang2010}, Joshi \textit{et al}.~\citep{Joshi2011} and~\citep{Lu2015}).
    \item \textbf{Observer tasks are not uniform} and may include biases~\citep{Joshi2011}. Tasks consist in most cases of scoring, but also some kind of description, and in some cases the inclusion of the image in 'favorites' or its sharing with others (e.g. Flickr.com). In many cases, it remains unclear whether participants are judging aesthetic beauty, technical quality, or both, as instructions are often vague or unspecified~\citep{Joshi2011}. For example, what is considered “interesting” in one platform (Flickr.com) may not be treated as being “aesthetically pleasing” in another (Photo.net).
    \item \textbf{The availability of highlights, descriptions,} comments, and ratings prior to voting fosters “preference groups” influenced by prestigious or expert users. Experts, for instance, show a stronger correlation between originality and quality~\citep{Hekkert1996} and tend to value challenging, emotionally provocative works more highly~\citep{Winston1992}.
    \item Some studies rely on binary distributions: Images are classified in only two categories (High/Low) to obtain high classification accuracy scores (see for example Datta \textit{et al}.,~\citep{Datta2006, Datta2010, Lu2015, Luo2011, Wang2019, Yan2006}. This represents an acknowledgment of the limitations of the methods and restricts the understanding one can derive from these results.
    \item \textbf{Some studies include arbitrarily chosen subsets}, e.g. certain “challenges”~\citep{Iigaya2021} or high-consensus images~\citep{Luo2011, Yan2006}. Others extract a large number of features (up to 1000) and calculate correlations without applying correction methods, such as the Bonferroni correction, to control for Type I errors (see for example Liu \& Wang,~\citep{Liu2017}).
    \item Some studies rely on a limited number of images or observers (see for example Bhattacharya \textit{et al}.~\citep{Bhattacharya2010},  Iigaya \textit{et al}.~\citep{Iigaya2021}, Jiang \textit{et al}~\citep{Jiang2010}, Liu \& Wang,~\citep{Liu2017}). Although there is no established consensus on the minimum required sample size, the high variability in responses—particularly in semantically rich datasets—indicates that robust analyses demand several thousand images and a substantial pool of observers to yield meaningful and generalizable results.
\end{itemize}

The purpose of outlining these limitations is not to undermine previous studies’ contributions, but to underscore the inherent challenges in applying methodologies centered on low- and mid-level feature extraction to a task that is predominantly influenced by high-level, cultural, and context-dependent factors. These points are not new and were thoroughly analyzed in a seminal work by Joshi \textit{et al}.~\citep{Joshi2011} which also examined other shortcomings of popular platforms at the time (Photo.net, DPChallenge.com and Terragalleria), e.g. their idiosyncratic distributions of average aesthetics scores, the presence of 'challenges', and their bias towards beauty, which are discussed below.

\subsection{Problems derived from other inherent database biases}
In most platforms, images are submitted under specific labels (“titles”, “themes” or “challenges”), a practice that sets a semantic background likely to affect subsequent aesthetic judgements. For instance, contributors to DPChallenge.com may evaluate an image by how close it fits to the “challenge” title, i.e., they might favor an image belonging to “Bad hair day” because it contains a pet with unruly hair or may prefer an image within “Through the eyes of a child” because it contains their favorite childhood toys. In other cases, the challenge’s winner decision may be decided on whether an image plays a humoristic twist on the issues of the day. In computational learning this is referred as “semantic gap”~\citep{Smeulders2000}, meaning the discrepancy between the information extracted from the visual data and the interpretation that observers make of the same data in a given context. Despite sharing certain structural features, all these databases contain large assemblies of objects with various semantic interpretations, depending on the situation, challenge, title, theme, etc. This contextual content acts as a strong confounding factor when it comes to predicting aesthetic valuations from the image visual content.

\begin{figure*}[htbp]
  \centering
  \includegraphics[width=0.8\textwidth]{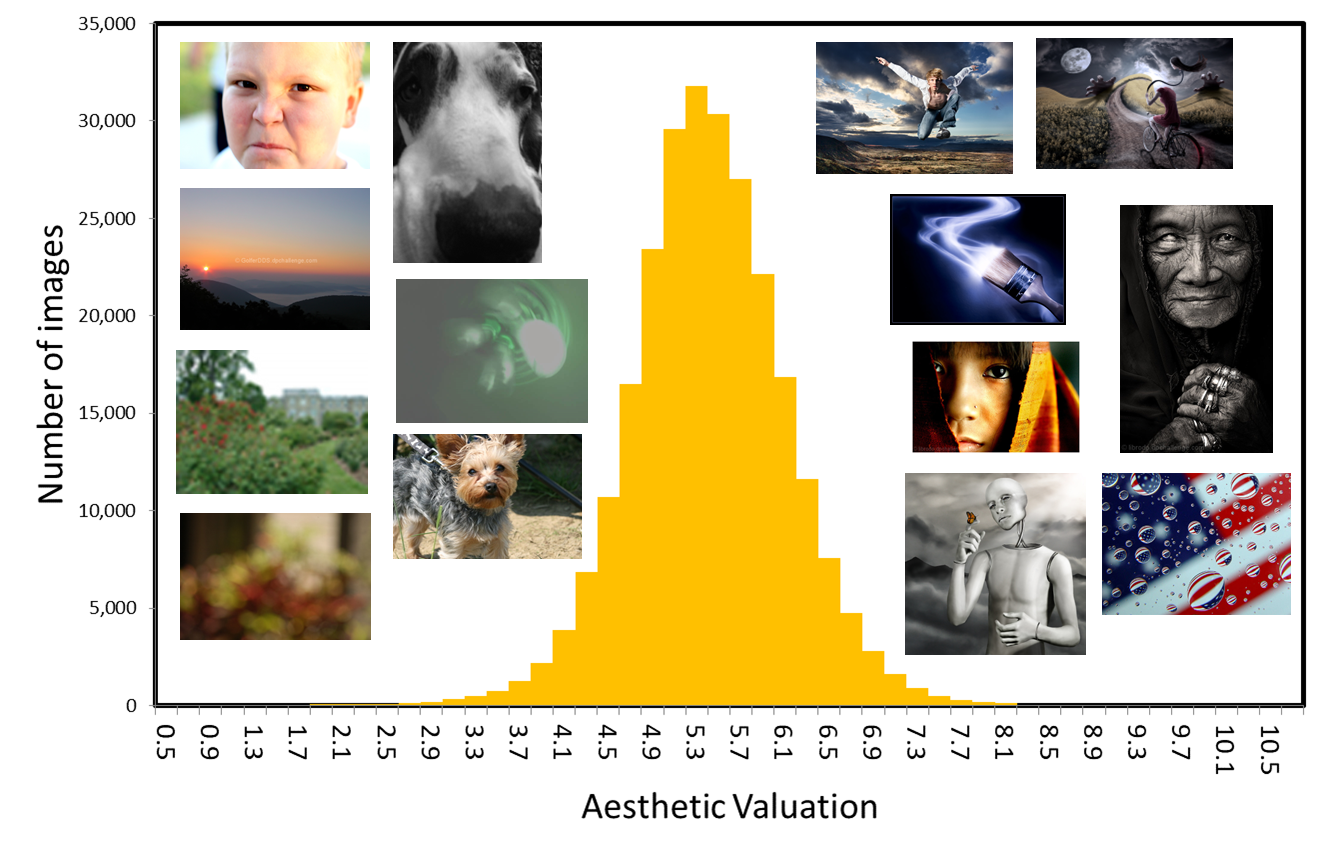}
  \caption{Distribution of aesthetic valuation responses for images in the most popular database in the literature, the Aesthetic Visual Analysis (AVA) database. The images on the right are the highest rated images of the database, similarly, the images on the left are the lowest rated.}
  \label{fig:Fig01}
\end{figure*}

Another major drawback of these databases is the small variability in the aesthetic dimension of the images contained and their arguable bias towards beauty~\citep{Joshi2011}. The low variability is illustrated by plotting the histogram of the number of images in each aesthetic valuation interval for the AVA dataset~\citep{Murray2012}. Figure~\ref{fig:Fig01} shows that images rated as “very ugly” (aesthetic valuation in the 0-1 range) or “very beautiful” (aesthetic valuation in the 9-10 range) are statistically rare, with most ratings in the range [4.5-6.5]. This pattern may be attributed to high inter-observer variability: extreme ratings are often offset by more moderate ones, and according to the law of large numbers, such variations tend to cancel each other out. As a result, the average aesthetic score for most images is drawn toward the center of the scale. This leads to a dataset with a limited dynamic range in the aesthetic dimension. In other words, despite the diversity of image content, there are very few images that elicit widespread agreement as either "very ugly" or "very beautiful." Consequently, the overall distribution of scores approximates a normal curve with a relatively narrow standard deviation.

Quantifying the overall bias towards beauty poses a challenge (we encountered no studies regarding the constraints of competition-based platforms); nonetheless, images with low aesthetic appeal are inherently less prevalent in contests and social networks, which serve as the primary sources for most databases. This is because participants typically refrain from submitting unappealing or ugly images to win or to showcase their work. To demonstrate this point, we added in Figure~\ref{fig:Fig01} the lowest rated images (left side) and the highest rated images (right side) of the AVA database. It could be stated that, lacking knowledge of the specific context under which the images were competing (i.e., the “theme” or “challenge”), even the lowest-rated images are, arguably, beautiful images.

The bias towards beauty combined with the limited dynamic range of the database, makes computational learning problematic. Indeed, any correlation analysis using a reduced range for one of the variables is likely to lead to spurious statistical outcomes, whether it overestimates, underestimates, or simply misses the correlation considering the full range of value for the variable. Another possible statistical trap when only considering a reduced range of values is to miss non-monotonic relationships between image metrics and aesthetic valuations. Some aesthetical valuations can, for example, depend on predictors such as complexity, following an inverted-U shape relationship~\citep{Berlyne1971}. Such a dependence would spuriously be reduced as a monotonic relationship when restricting the range of values for the predictor.

\subsection{The Minimum Semantic Content (MSC) database: a reductionist approach to empirical aesthetics}

In this study we build a novel database, the Minimum Semantic Content (MSC) database, which tackles the three issues that hamper a systematic analysis of the links between image features and aesthetic judgements, namely: (1) the entanglement between cognitive and perceptual valuation, (2) the systematic bias towards beautiful images, and (3) the lack of extreme examples. For this purpose, we first gather a large set of images that only contain objects from the natural environment (landscapes, trees, rocks, mountains, water, etc.) and therefore could be considered as lacking the elements that elicit the strongest cognitive and/or emotional responses, or, at least, to make semantic content uniform. Indeed, using “ugly” or synonyms as a keyword for retrieving internet images leads to outputs with strong emotional and cognitive content. For example, a quick Google search with the keywords “ugly images” produces a collection of grotesquely deformed human faces, suffering or diseased animals, shockingly violent images, feces, etc. This might be a consequence of humans evolving various emotional mechanisms such as fear, as a response to threats (e.g., predators, heights, starvation) or disgust (including sexual and moral disgust) as a response to risks and disease~\citep{Klebl2022}. Moreover, studies on nature and urban imagery~\citep{Felisberti2022} suggest that the experience of ugliness is an independent experience not only entangled with the negation of beauty, but also able to coexist with it.

The new database we propose, characterized by a limited (arguably minimal) level of semantic content, enables us to explore the impact of image features on aesthetic judgement, free from the predominant influences of cognitive factors. In this regard, we adopt a reductionist methodology where the emotional and cognitive aspects of aesthetic judgments are held constant, facilitating the examination of the independent "perceptual" component (see Redies~\citep{Redies2015}). Simultaneously, we mitigate the influence of multiple "themes" or "challenges" that could have contributed to the properties of the distribution in Figure~\ref{fig:Fig01}.

To address the bias towards beauty and lack of extreme examples in existing databases, we developed a digital image editing software to modify images. Instead of retrieving ugly images from the web, we provide observers with tools to change low-, mid-, and high-level image features and explicitly ask them to transform the original minimum semantic content images into ugly images. This novel approach allowed us to also tackle the ubiquitous ugliness/emotion entanglement that we found in internet searches, where images with high semantic / negative emotional content are likely be associated with low aesthetic value.

It is important to note that we explicitly avoided the inclusion of any type of artwork in our database. There is extensive literature showing the impact of contextual information on the aesthetic judgment of artworks~\citep{Darda2023}, which we wanted to minimize. Similarly, we did not include databases of art or art-related images in our comparative analysis. We avoided art representations, as our goal was to create a controlled experimental setting that isolates perceptually-driven aesthetic decisions from those influenced by cognitive or semantic factors. By minimizing semantic content, the MSC dataset enables a focused examination of perceptual factors underlying aesthetic judgments without the confounding effects introduced by recognizable or culturally loaded content.

The contributions of this article can be summarized as follows: (1) we provide a database of aesthetically valuated images (originally based on images of natural scenes) that were balanced to counter biases towards beauty and whose semantic content was significantly reduced, (2) we augment studies of the relationship between aesthetic valuations and several low- and mid-level image features to the full spectrum of aesthetics valuations, and (3) we systematically describe the changes made by observers to remove biases towards beauty.

\section{Method}
\label{sec:methods}

\subsection{Dataset creation overview}

We took the following steps to generate our image database:
\begin{itemize}[leftmargin=*, itemsep=0pt, topsep=0pt]
\item \textbf{Collect copyright free images.} We collected a large amount of images exclusively from copyright-free or creative commons online databases~\citep{BurningWell2004, Flickr2003, GoodFreePhotos2012, Morguefile1996, PdPhoto2003, PhotosPublic2010} and own sources.
\item \textbf{Remove semantics and emotions.} We excluded images containing rich semantic content, i.e., those containing people, animals, or human-made objects from the original database. We further eliminated obvious examples of “postcard landscapes” or “holiday brochures’ landscapes” that may also elicit emotional responses. The resulting 5,684 images were later cropped to eliminate frames, colored borders, etc. and then resized to make them easy to process and distribute. The final sizes ranged between 293-800 pixels in height and 334-1200 pixels in width. We called this collection the “original” database.
\item \textbf{Create a subset of modified images.} We asked a small group of observers to manipulate the spatiochromatic characteristics of some of the original images to create either ugly or beautiful scenes (see details below). We called these new images either “beautified” or “uglified”, in agreement with the instructions given to their creators.
\item \textbf{Augment the database.} We first added the modified (both beautified and uglified) images to the original images. Next, we created further images by recording the uglification manipulations and randomly applying them to some of the original images. We called these new images “auto-uglified” following the previous naming convention.
\item \textbf{Obtain aesthetic valuations.} We obtained valuations of the whole database from a large number of non-expert observers using a crowdsourcing paradigm.
In the end, we obtained 10,426 copyright-free images with minimum semantic content. This included a sizable proportion of scenes that were either beautified (8.8\%), uglified (8.4\%), or automatically manipulated (auto-uglified- 28.3\%). The image manipulations are explained below.
\end{itemize}

\begin{figure*}[htbp]
  \centering
  \includegraphics[width=0.8\textwidth]{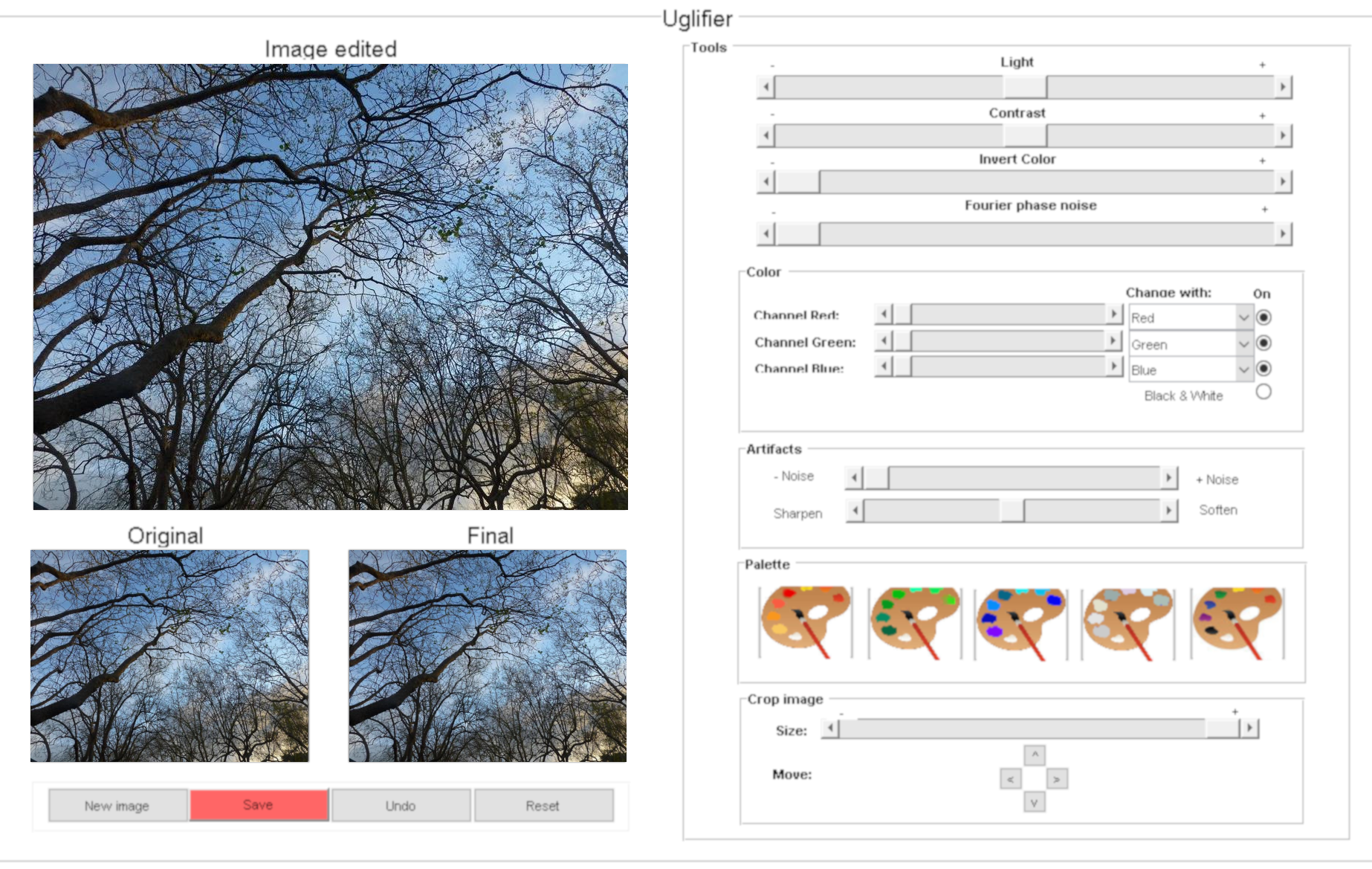}
  \caption{Screenshot of the “\textit{uglifier}” user interface. Images were presented to the left and users applied the different manipulations by moving the sliders. The software allowed them to undo the previous manipulations, restart (perform a reset), save, etc., and always keep the original image visible.}
  \label{fig:Fig02}
\end{figure*}

\subsection{The “\textit{Uglifier}”, an ad-hoc image manipulation software}

To compensate for the bias towards beautiful images in the original database, we made a computational tool that allowed observers to manipulate the low-level features of images (see Figure~\ref{fig:Fig02}). This tool (called the “Uglifier”) was designed to manipulate features described as important for aesthetic valuation in previous research with a user-friendly interface. Through a series of trials with naïve observers, we identified effective and intuitive manipulations, avoiding those that were overly complex or time-intensive to prevent participant fatigue or predictable responses. These manipulations were chosen not only for their ability to adjust key visual features—such as color, contrast, and spatial frequencies—but also to foster a broader exploration of aesthetic preferences.

We asked 40 observers (university students, 30 male and 10 female) to manipulate 1,791 images randomly selected from the original database (approximately 45 images each). They were instructed to “beautify” 919 of them (i.e., to make them as beautiful as possible) and “uglify” 872 of them (i.e., to make them as ugly as possible). As these observers knew the purpose of the manipulations, and by opposition to the naïve observers (see below), we call them informed observers.

We recorded the manipulations performed when informed observers uglified images and generated another subset of 2,951 images by applying them randomly to the same number of original images. The idea here was to augment the data destined to counter the beauty bias since the manipulated images were expected to score low in the aesthetics axis. Table~\ref{tab:Table02} shows a list and a description of the image manipulation modules available to users of the uglifier algorithm.

\begin{figure*}[htbp]
  \centering
  \includegraphics[width=0.8\textwidth]{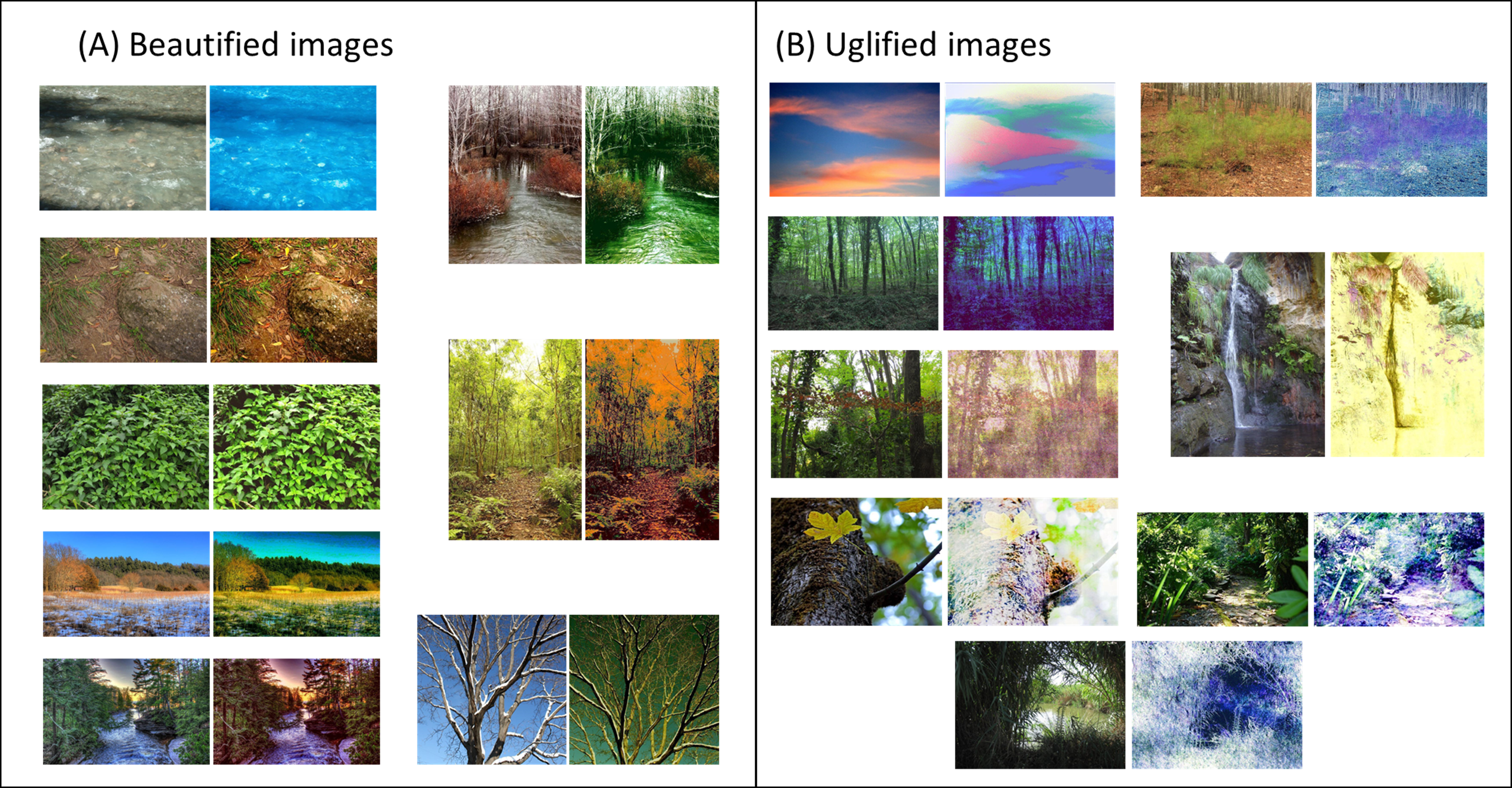}
  \caption{Screenshot of the “\textit{uglifier}” user interface. Images were presented to the left and users applied the different manipulations by moving the sliders. The software allowed them to undo the previous manipulations, restart (perform a reset), save, etc., and always keep the original image visible.}
  \label{fig:Fig03}
\end{figure*}

Figure~\ref{fig:Fig03} shows examples of images manipulated using the uglifier. Informed observers could preview the results of their work and undo them if needed. The software also recorded all the steps from the beginning until the final product was saved. We later used these recordings to automatically generate new images (auto-uglified images). After adding the modified images to the original database, we ended up with a total of 10,426 images that were later crowdsourced to establish their average aesthetic valuations.

\begin{table*}[htb!]
\centering
\caption{Description of the different image manipulation modules of the Uglifier algorithm available to users.}
\label{tab:Table02}
\renewcommand{\arraystretch}{1.2} % add some padding between lines
\begin{tabular}{p{5cm}p{9cm}}
\toprule
\textbf{Module name} & \textbf{Effect on the image} \\
\midrule
Lightness slider & Overall lightness increase/decrease. \\
Contrast slider & Overall contrast increase/decrease. \\
Color inversion slider & Image color inversion. Switches between opponent colors (red/green, yellow/blue and light/dark). \\
Fourier phase noise slider & Introduces increasing amounts of randomization to the phase component of the Fourier representation of the image. Destroys the image content while keeping the original second-order Fourier statistics. \\
Color manipulation (contains three separate RGB channel sliders) & RGB relative contribution is increased/decreased. It can also exchange, disable, or equalize RGB channels (i.e., convert image to grayscale). \\
Artifacts (contains a random noise slider and a blur/sharpen slider) & Adds random noise to the image, Blurs the image or performs edge enhancement using a circularly symmetric Gaussian filter. \\
Palette exchange & Exchanges the color palette with that of an external image which could be either reddish, bluish, greenish, or whitish (see Pitie \textit{et al}., 2007). \\
Cropping and positioning slider & Allows to select a portion of the original image and discard the rest. \\
\bottomrule
\end{tabular}
\end{table*}

\subsection{Crowdsourcing}

\begin{itemize}[leftmargin=*, itemsep=0pt, topsep=0pt]
\item \textbf{Basic experiment outline.} Although a portion of the images were created with the specific aim of making them as beautiful or ugly as possible, we did not rely on those judgments for our database valuation. Instead, we used a third-party crowdsourcing paradigm (implemented by \textit{Knowxel Crowdmobile S.L.} - Cerdanyola, Barcelona) to obtain aesthetic valuations from many (more than 10,000) naïve observers over the internet. The crowdsourcing company also had algorithms to flag and discard observers who pressed buttons randomly or did not fully commit. The experiment consisted of two stages. The first was an introductory/training stage where observers received an explanation of the task and were presented (in random order) with a subset of 10 images created to be either very beautiful or very ugly. They were asked to rate them on a Likert scale from 1 to 5, (1 = very ugly and 5 = very beautiful). The purpose of the introductory stage was to expose naïve observers to the full dynamic range of the database and the results were later discarded. The second stage was the “actual” experiment. Observers rated a minimum of 14 random images from the image database. Our paradigm allowed them to change their vote in case they selected the wrong option by mistake.
\item \textbf{Pilot run.} To ensure that the database was not biased, we made two runs of the crowdsourcing experiment. The first run, a pilot, included 12,909 images of which 2,000 were modified by informed observers (1,000 uglified and 1,000 beautified) with the Uglifier and 1,000 randomly modified as described below. The results showed a strong bias towards the beautiful side (almost 60\% of the database was rated between 4 and 5), which was not surprising, given that most images came from online sources. This prompted us to make a second run with a less biased selection of images, randomly discarding some beautiful examples and adding more randomly modified ones. We kept 919 beautified images, 872 uglified images, 2,951 auto-uglified images and 5,684 unmodified (original) images, resulting in a database consisting of 10,426 images.

\item \textbf{Final experiment.} The second run of the experiment was just a repetition of the crowdsourcing with this new, unbiased MSC database from scratch. In the end, we obtained 100 valuations for each of the 10,426 images. The total number of naïve observers was larger than 10,000, although we had no means to control how many times a single person participated in the experiments, their gender or physical location (they were approximately evenly distributed among Europe, Asia, and the Americas).
\end{itemize}

\subsection{Obtaining aesthetic valuations from crowdsourcing results}
Although most of the aesthetic valuations followed a typical normal (Gaussian) distribution, many did not fit this profile. For our purposes, we wanted to extract a single valuation from each distribution. This is usually obtained by selecting the largest bin of the histogram or computing the statistical normal of all valuations instead. We also wanted to calculate correlations between valuations and several image features and for this, it was more convenient to use the information in the histograms to break the discretization of the results, converting them into a continuous scale. Arguably, using the average rating value also breaks the discretization, but this will severely compress the dynamic range of the results and few images will obtain extreme valuations.

\begin{figure*}[htbp]
  \centering
  \includegraphics[width=0.8\textwidth]{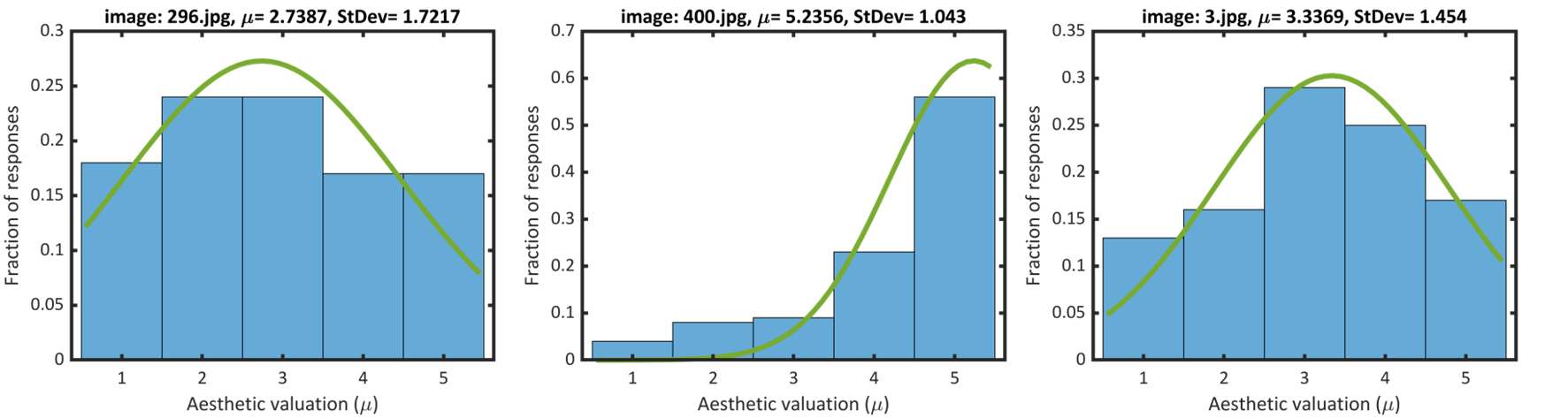}
  \caption{Examples of truncated histogram fittings of valuation results for three images of the database. The green lines represent the truncated histogram fittings. Mean ($\mu$) and SD of the distributions are included in each plot.}
  \label{fig:Fig04}
\end{figure*}

After a visual inspection of the data, we decided to fit a truncated Gaussian (see Figure~\ref{fig:Fig04}) to each histogram. This provided two statistical descriptors: the mean ($\mu$) and the SD of each valuation distribution. For the fittings, we assumed that the first bin corresponded to valuations between 0.5 to 1.5, the second bin to valuations between 1.5 and 2.5, and so on. Our fitting code was customized from Ryabov’s code~\citep{Ryabov2022} with the following constraints: (a) the value of $\mu$ was restricted between 0.5 and 5.5, which is the full range of the bars and (b) the SD of the Gaussian distribution was not allowed to have values below 0.5 (half of a histogram bar). Using these constraints, each image of the database was assigned a $\mu$ value in the range [0.5, 5.5] and a SD in the range [0.5, 1.9].

Figure~\ref{fig:Fig04} shows three examples of histogram distributions and their corresponding truncated Gaussian fittings. Appendix Figure~\ref{fig:FigA1} confirms that computing the average or the median severely restricts the dynamic range of valuation responses (picture A) and that $\mu$ is highly correlated to results obtained using the statistical median (r² = 0.89), the peak value (r² = 0.86) and the average (r² = 0.96).

\begin{figure*}[htbp]
  \centering
  \includegraphics[width=0.8\textwidth]{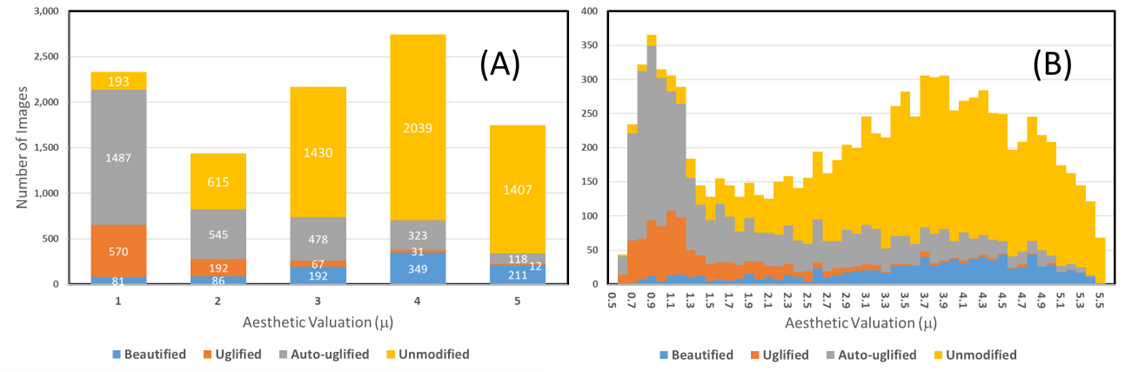}
  \caption{Distribution of aesthetic valuations ($\mu$) in our database. (A) first picture shows the total number of images for each of the five Likert scale categories and types of images; (B) second picture shows the same distribution with higher bin resolution. The blue sections correspond to beautified images, the orange sections to uglified images, the grey sections to auto-uglified images and the yellow sections to unmodified images.}
  \label{fig:Fig05}
\end{figure*}

\subsection{Distribution of aesthetic valuations}
Figure~\ref{fig:Fig05} shows the distribution of mean aesthetic valuations ($\mu$) in our database, calculated using the method described above. The plots show that most of the uglified images had low ratings, most of the beautified images had medium and high ratings and a large number of the auto-uglified images were also rated lowly by our crowdsourcing observers. Figure~\ref{fig:Fig05} also shows that we managed to remove the bias towards high ratings, creating a more uniform distribution, which was one of our main objectives.

Table~\ref{tab:Table03} shows a comparison between the MSC database and some of the most popular databases used in computational aesthetics, summarizing the number of images, source, ratings per image, etc.

Once we had assigned a mean rating value ($\mu$) and a SD to each image in the database, we were ready to test whether the removal of biases and high-level semantic content would facilitate the extraction of features that contributed significantly to the aesthetic valuation decision and were hidden in other databases.

\begin{table*}[htb!]
\centering
\caption{Comparison between our database and the most popular databases used in computational aesthetics in terms of number of images, number of ratings per image, etc.}
\label{tab:Table03}
\renewcommand{\arraystretch}{1.2} % add some padding between lines
\begin{tabular}{p{2cm}p{2.5cm}p{2.5cm}p{2.5cm}p{2.5cm}}
\toprule
\textbf{Database name} & \textbf{AVA} & \textbf{DPChallenge} & \textbf{Photo.net} & \textbf{MSC database (this study)} \\
\midrule
Nr. of images & 250,000 & 16,509 & 3,581 & 10,426 \\
Platform & DPChallenge.com & DPChallenge.com & photo.net & Flickr, PDPhoto, Morguefile, own images, etc. \\
Ratings/image & 210 & 205 & 12 & 100 \\
Characteristics & Semantic and photographic style annotations &  &  Semantic and photographic style annotations & Some images purposely uglified or beautified \\
Rating Score & 1 to 10 & 1 to 10 & 1 to 7 & 1 to 5 \\
Copyright & Copyrighted & Copyrighted & Copyrighted & Open Commons \\
\bottomrule
\end{tabular}
\end{table*}

\subsection{Aesthetic assessment metrics}
Many image characteristics have been related to beauty, mainly from neuroscience and computational points of view~\citep{Brachmann2017, Datta2006}. Since one of the objectives of this work is to showcase the importance of removing biases and semantics from the stimulus database, we selected a few good candidates features whose extraction is relatively easy to implement. Table~\ref{tab:Table04}, at the end of the section, provides a brief description of the metrics evaluated. For a justification, please refer to the discussion section below.

\subsection{Basic, low-level features}

\begin{itemize}[leftmargin=*, itemsep=0pt, topsep=0pt]
\item \textbf{Chromatic and achromatic multiscale contrast.} We calculated contrast by filtering each of the chromatically opponent channels (“a” and “b”) of CIELab space and the “L” channel with differences of Gaussians (DoG) at various spatial scales. Images were converted from RGB to CIELab using MATLAB’s $rgb2lab$ function. One of the components of the DoG was a circularly symmetric Gaussian whose initial standard deviation (SD) was half the size of the image diagonal and was decreased in octaves (1/2, 1/4, 1/8, etc.) to simulate the spatial scales present in the visual cortex until a minimum value of 4 pixels was reached. The other component was a negative Gaussian with a SD twice as large. The final contrast was found by filtering and averaging the results across all pixels and spatial scales. Three values were obtained for each picture using equations~\eqref{eq:contrast1} to~\eqref{eq:contrast3} as described in the appendix.
\item \textbf{Global contrast factor (GCF).} This metric was obtained from the literature~\citep{Matkovic2005} and computes contrasts at various resolution levels in order to obtain a single measure of overall contrast. Weighting factors were obtained from psychophysical experiments in support of the authors’ claim that GCF better corresponds to human perception. A MATLAB version of the code was obtained from GitHub~\citep{Povoa2013}.
\item \textbf{Statistical measure of colorfulness (CIELab gamut expansion).} We computed the simplest measure of colorfulness, i.e., the average distance of pixels to the white locus in CIELab color space. Images were converted from to CIELab using MATLAB’s rgb2lab function using the default white point (‘D65’).
\item \textbf{Overall perceived colorfulness.} This metric was obtained from the literature~\citep{Hasler2003} and quantifies the (overall) perceived colorfulness of natural images from the distribution of the pixels in CIELab space using a linear combination of quantities (the SD along the a and b axis, the trigonometric length of the SD in ab space, the distance from the center of gravity in ab space to the neutral axis, etc.). The authors show that it correlates strongly with their psychophysical results. The MATLAB implementation was obtained from the GitHub repository~\citep{Povoa2015}.
\item \textbf{Saturation (from the HSV color model).} This simple metric converts an RGB image to hue, saturation, and value (HSV) using MATLAB’s $rgb2hsv$ function, and preserves the average saturation “S”.
\item \textbf{Mean lightness value.} Computes the average CIELab lightness “L” for the whole image.
\item \textbf{Focus/Blur metric.} This metric was also obtained from the literature~\citep{Thelen2009}. It measures the relative degree of focus of an image using a diagonal Laplacian and was developed to obtain shape from focus for holographic reconstructions. Its effectiveness was compared to other up-to-date algorithms~\citep{Pertuz2013}. The code was obtained from the MATLAB repository~\citep{Pertuz2022}.
\item \textbf{Fourier amplitude slope ($\alpha$).} This metric was chosen because of its links to physiological properties of the visual cortex~\citep{Field1987, Knill1990, Parraga2000}, heightened sensitivity or preference toward natural spectra and visual discomfort~\citep{Penacchio2023, Penacchio2015}. We calculated the Fourier amplitude slope in three steps: (a) convert all images to grayscales using MATLAB’s $rgb2gray$ function; (b) average Fourier amplitude over all orientations for each possible spatial frequency and, (c) plot the log of the resulting Fourier amplitude vs the logarithm of the spatial frequency and fit a line to the results. The slope of this line (a negative number) is the Fourier amplitude slope ($\alpha$). A higher negative value represents images with more Fourier energy in the low-spatial-frequencies (i.e., they contain large areas with relatively constant luminance), and a less negative value represents images with more Fourier energy in the high-spatial-frequencies (i.e., contain large areas with fine textures). The traditional way to measure $\alpha$ entails computing a value for every single spatial frequency component in Fourier space, averaging all orientations. When we plot these average Fourier amplitude values against spatial frequency in log-log space, most of the points are concentrated near the high end of the spectra, dominating the slope calculation. This problem becomes particularly bad in some images, e.g., those containing high energy noise. To counter the log-log space bias, we implemented another way of measuring $\alpha$: instead of averaging Fourier amplitude over all orientations for each possible spatial frequency, we averaged it for all orientations within spatial frequency “bands”. To achieve a logarithmically uniform distribution across frequencies, the widths of the bands were designed to increment in octaves. This improves line-fitting in log-log space, effectively reducing the impact of high spatial frequencies on the calculation of $\alpha$~\citep{Brelstaff1995}. We measured the slope of the amplitude spectrum using both methods (“traditional” and “unbiased”). It is important to notice that a steeper $\alpha$ correlates with, but is not the same as blurring, since it involves changes at both ends of the Fourier spectrum.
\item \textbf{Image complexity.} This consists of two metrics developed by Groen \textit{et al}~\citep{Groen2013} to describe the complexity of an image in terms of its contrast distribution and scene fragmentation using summary statistics obtained from the spatial pooling of responses of neurons in the visual pathway of the brain~\citep{Scholte2009}. The first metric, called Contrast Energy (CE), describes the scale of contrast in an image. The second metric, called Spatial Coherence (SC), describes the degree of correlation between contrast elements in the image. Both metrics convolve the image with multiscale filters (smaller sizes for CE than for SC) and for each image location, a single filter response is selected and pooled across a selection of the visual field (for CE responses were averaged and for SC a coefficient of variation was obtained). These were later averaged across the three color-opponent representations resulting in one CE and SC value per image. In their article, Groen \textit{et al}. show that CE describes the scale of the contrast distribution (varies with the distribution of local contrasts strengths) and SC describes the shape of the contrast distribution (varies with the amount of scene fragmentation/clutter). They also show that these metrics (SC in particular) are related to single-trial event-related potentials (sERPs) obtained during a naturalness rating task. The code to compute CE and SC was kindly provided by Iris Groen. We calculated both metrics using the default parameters for their code.

\subsection{Higher level features (machine-learning-based)}

We also explored three extra features, related to higher brain functions, which were obtained using machine learning techniques:

\item \textbf{Symmetry.} We tried the symmetry metric developed by Brachmann and Redies~\citep{Brachmann2016} who used a modification of CaffeNet~\citep{Krizhevsky2017}, a Convolutional Neural Network (CNN) that was trained on the ImageNet database. They validated their algorithm on music album covers, which were rated according to their symmetry by human observers. The algorithm provides several options, and we chose the Left-Right and Top-Down variants. The authors claim that their model of symmetry closely resembles human perception of symmetry in CD music album covers. A Python version of the code was obtained from OSF~\citep{Redies2019}.
\item \textbf{Mean 3D depth.} This monocular metric~\citep{Ranftl2021} tries to quantify the mean depth of the image using dense prediction transformers, an architecture that leverages vision transformers instead of convolutional networks. Images in the database were reduced to have a maximum side size of 150 pixels to save computer processing power. The PyTorch-converted models were obtained from GitHub~\citep{Ranftl2012}.
\end{itemize}

\begin{table*}[htb!]
\centering
\caption{Brief description of the aesthetic assessment metrics explored}
\label{tab:Table04}
\renewcommand{\arraystretch}{1.2} % add some padding between lines
\begin{tabular}{p{6.5cm}p{8.5cm}}
\toprule
\textbf{Metric} & \textbf{Description} \\
\midrule
&  \textbf{ Basic, low-level features} \\
\cline{2-2}
Multiscale Contrast (C\textsubscript{Lab}) & Contrast calculated from the three CIELab channels. \\
Multiscale Contrast (C\textsubscript{L}) & Contrast calculated from the CIELab “L” channel only. \\
Multiscale Contrast (C\textsubscript{ab}) & Contrast calculated from the CIELab “a” and “b” channels. \\
Global Contrast Factor~\citep{Matkovic2005} & Single measure of overall contrast, psychophysically tested. \\
Statistical Colorfulness (CIELab Gamut Exp.) & Average distance of pixels to the white locus in CIELab. \\
Perceived Colorfulness~\citep{Hasler2003} & Linear combination of statistical quantities in CIELab, psychophysically tested. \\
Saturation (HSV model) & Simple saturation metric from MATLAB. \\
Mean Lightness (CIELab “L”) & Average CIELab “L” value for the whole image. \\
Focus/Blur metric~\citep{Pertuz2013} & Relative degree of focus of an image using a diagonal Laplacian. \\
Fourier amplitude slope (\(\alpha\)) (traditional) & Fourier amplitude slope calculated for all spatial frequencies. \\
Fourier amplitude slope (\(\alpha\)) (unbiased) & Fourier amplitude slope calculated over averaged spatial frequency “bands”. \\
Image Complexity~\citep{Groen2013} & Metrics describing the complexity of an image in terms of its contrast energy distribution and spatial coherence (scene fragmentation). \\
\cline{2-2}
& \textbf{Higher level features (machine-learning)} \\
\cline{2-2}
L-R and U-D Symmetry metric~\citep{Brachmann2016} & Left-Right and Up-Down variants of a symmetry metric obtained by a CNN trained on the ImageNet database. Validated by human observers on CD album covers. \\
Mean 3D-depth~\citep{Ranftl2021} & Quantifies the mean depth of the image using dense prediction transformers. \\
\bottomrule
\end{tabular}
\end{table*}

\subsection{Results}
We applied the above metrics to the MSC database and obtained correlations with the mean ratings $\mu$. Given that auto-uglified images were the product of applying the uglification in a random manner and were mostly considered ugly in the crowdsourcing (see Figure~\ref{fig:Fig05}), we combined them with the uglified for our analysis.
Table~\ref{tab:Table05} summarizes the correlation results including an extra column for those of AVA~\citep{Murray2012}, which is arguably one of the best documented and popular databases available today. To compare MSC to AVA we split our correlation plots into two categories: all images and unmodified images only. The first allows us to visualize the contribution of modified images to the results and the second underlines how, by simply removing semantics, we can unveil correlations that are hidden in other databases. In summary, the difference between the first and third columns in Table~\ref{tab:Table05} highlights the impact of removing the bias towards beautiful imagery, and the difference between the third and fifth columns highlights the confounding role of semantics in aesthetics research.

Detailed results for each metric applied to MSC are presented below. We separated the MSC into three categories of images: (a) uglified \& auto-uglified, (b) unmodified and (c) beautified. Because of the large number of datapoints, we highlighted prominent clusters using colored scatterplots from the MATLAB repository~\citep{Lukas2023}. Correlations (r) refer to Pearson’s r and their corresponding p-values. Also because of the large number of datapoints, many p-values are extremely small (e.g., p< 10-10) and are quoted as “p<0.001”. The exact p-values are in Table~\ref{tab:Table05}.

\begin{table*}[htb!]
\centering
\caption{Correlations between the values obtained for the different features and aesthetic valuations. The table shows Pearson’s $r$ and their corresponding $p$-values for all images (first column) and unmodified images (second column) in our database. For comparison, we added a third column with the same results obtained for the AVA database. Since original AVA valuations are in a 1 to 10 scale, they were reduced to 1 to 5 by adding the votes of pairs of consecutive bins. The mean ($\mu$) was obtained by applying the same methods described before (see Figure~\ref{fig:Fig04}).}
\label{tab:Table05}
\renewcommand{\arraystretch}{1.2} % add some padding between lines
\begin{tabular} {p{6.5cm} p{1cm} p{1cm} p{1cm} p{1cm} p{1cm} p{1cm} }
\hline
\textbf{Metric} & \multicolumn{2}{c}{\textbf{\parbox{2.5cm}{\centering MSC database \\ \centering (all)}}} & \multicolumn{2}{c}{\textbf{\parbox{2.5cm}{\centering MSC database \\ \centering (unmodified)}}} & \multicolumn{2}{c}{\textbf{AVA database}} \\
\hline
 & $r$ & $p$ & $r$ & $p$ & $r$ & $p$ \\
\cline{2-7}
Multiscale Contrast ($C_{Lab}$) & 0.513 & 0 & 0.499 & 0 & -0.002 & 0.216 \\
Multiscale Contrast ($C_L$) & 0.498 & 0 & 0.332 & $10^{-146}$ & 0.008 & $10^{-4}$ \\
Multiscale Contrast ($C_{ab}$) & 0.272 & $10^{-176}$ & 0.406 & $10^{-224}$ & 0.017 & $10^{-17}$ \\
Global Contrast Factor~\citep{Matkovic2005} & 0.220 & $10^{-104}$ & -0.004 & 0.7 & 0.004 & 0.03 \\
Statistical Colorfulness (CIELab Gamut Exp.) & -0.224 & $10^{-119}$ & 0.200 & $10^{-52}$ & -0.015 & $10^{-15}$ \\
Perceived Colorfulness~\citep{Hasler2003} & -0.029 & 0.003 & 0.356 & $10^{-170}$ & -0.008 & $10^{-5}$ \\
Saturation (HSV model) & -0.088 & $10^{-19}$ & 0.256 & $10^{-87}$ & 0.002 & 0.3 \\
Mean Lightness (CIELab “L”) & 0.135 & $10^{-43}$ & -0.017 & 0.2 & -0.034 & $10^{-66}$ \\
Focus\& Blur metric~\citep{Pertuz2013} & -0.037 & 0.0001 & 0.025 & 0.05 & 0.064 & $10^{-228}$ \\
Fourier $\alpha$ slope (traditional) & -0.216 & $10^{-110}$ & 0.155 & $10^{-32}$ & 0.227 & 0 \\
Fourier $\alpha$ slope (unbiased) & -0.209 & $10^{-103}$ & -0.117 & $10^{-19}$ & 0.174 & 0 \\
Image Complexity~\citep{Groen2013} & 0.171 & 0 & -0.002 & 0.877 & 0.043 & $10^{-105}$ \\
L-R Symmetry metric~\citep{Brachmann2016} & -0.344 & $10^{-288}$ & -0.270 & $10^{-95}$ & -0.026 & $10^{-42}$ \\
U-D Symmetry metric~\citep{Brachmann2016} & -0.361 & $10^{-317}$ & -0.337 & $10^{-150}$ & -0.008 & 0 \\
Mean 3D-depth~\citep{Ranftl2021} & -0.188 & $10^{-84}$ & -0.271 & $10^{-96}$ & -0.085 & 0 \\
\hline
\end{tabular}
\end{table*}

\subsection{Chromatic and achromatic multiscale contrast}

\begin{figure*}[ht]
    \centering
    \includegraphics[width=0.8\textwidth]{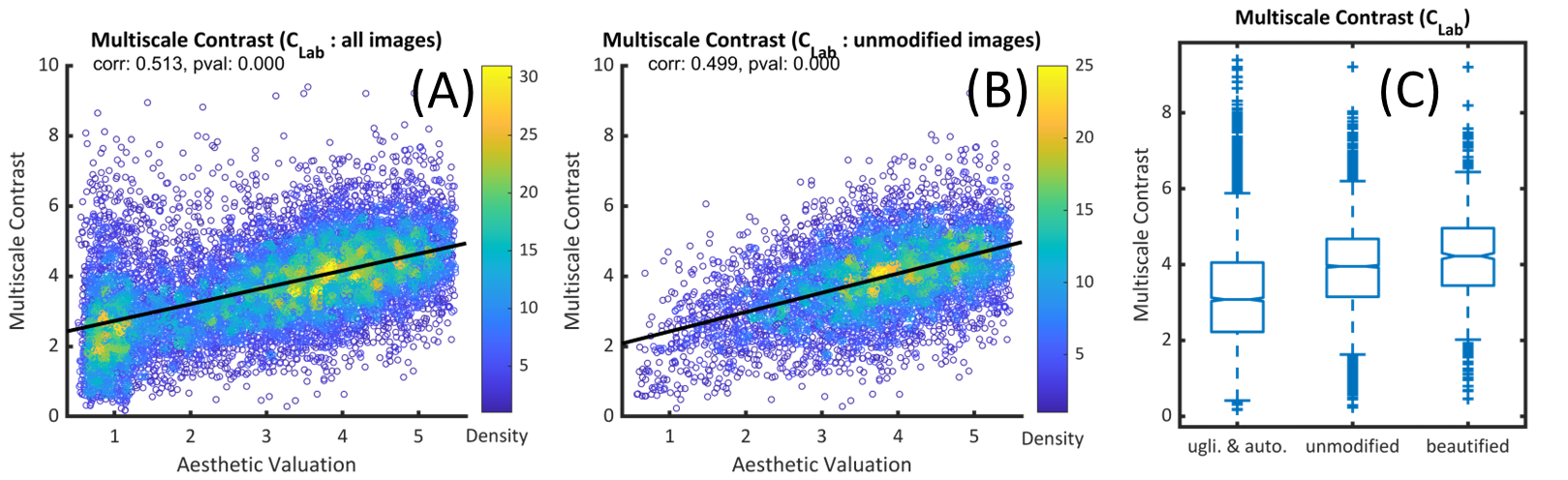}
    \caption{Correlation between the multiscale contrast ($C_{Lab}$) and aesthetic valuation ($\mu$). (A) Scatterplot for all images. The color shows the density of points, with a yellower color corresponding to a higher density and a bluer color to a lower density. (B) Same for unmodified images (i.e., excluding uglified, beautified and auto-uglified). (C) Metric values for the three groups of images considered (uglified and auto-uglified, unmodified, and beautified). Boxplots show median value, 25th and 75th percentiles (lower and upper hinges), the lowest measured values within Q1 (first quantile) and 1.5 x Q1 (lower whisker) and the highest observed value within Q3 (third quantile) and 1.5 x Q3 (upper whisker).}
    \label{fig:Fig06}
\end{figure*}

Figure~\ref{fig:Fig06} shows the results obtained from applying the multiscale contrast ($C_{Lab}$) metric to our database and plotting it against the mean aesthetic valuation ($\mu$) as described in the methods section. Picture A considers all the images, picture B considers unmodified images only, and picture C shows the boxplots of this metric for our three groups: the uglified \& auto-uglified, the unmodified and the beautified images. Both ANOVA and Tukey–Kramer tests confirmed that the means of the three boxplots (3.228, 3.907 and 4.212) were significantly different (pairwise $p<0.001$).

Picture A also shows the correlation ($r=0.513$, $p<0.001$) between $C_{Lab}$ and $\mu$ with a concentration of points to the left of the plot (low aesthetic valuations), corresponding mostly to uglified images (see the histogram in Figure~\ref{fig:Fig05}). Picture B shows the same ($r=0.499$, $p<0.001$) for unmodified images only. We also analyzed this metric considering $C_L$ (lightness only) and $C_{ab}$ (chromatic channels only) with qualitatively similar results (see Figure~\ref{fig:FigA2} and Figure~\ref{fig:FigA3} in the appendix). The boxplots in Picture C show that, when creating ugly images, informed observers follow the same trend already present in unmodified images (picture B), lowering the contrast and increasing its correlation with $\mu$.

\subsection{Global Contrast Factor (GCF)}

We included the Global Contrast Factor (GCF) metric of Matković \textit{et al}. to provide a perceptual correlate to the purely statistical, DoG-based multiscale contrast.

\begin{figure*}[ht]
    \centering
    \includegraphics[width=0.8\textwidth]{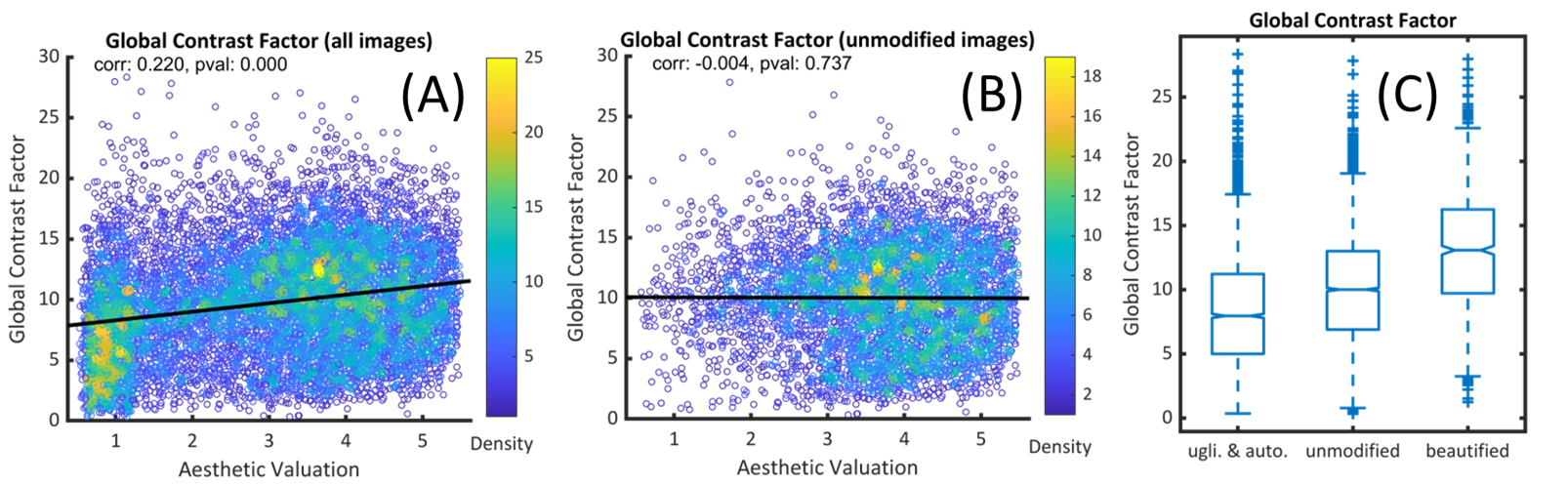}
    \caption{Results from the Global Contrast Factor (GFC) defined by Matković \textit{et al}. (A) Scatterplot for all images. (B) Same for the unmodified images. (C) Metric values for the three groups of images considered. Boxplots as in Figure~\ref{fig:Fig06}.}
    \label{fig:Fig07}
\end{figure*}

Figure~\ref{fig:Fig07} shows the correspondence between GCF and mean aesthetic valuation ($\mu$). Pictures A and B show GCF vs $\mu$ for all images and for unmodified images respectively. Picture C shows the boxplots of this metric for the uglified \& auto-uglified, the unmodified and the beautified images. Both ANOVA and Tukey–Kramer tests confirmed that the means of the three boxplots (8.425, 10.02 and 13.104) were significantly different (pairwise $p<0.001$).

Again, picture A shows a correlation ($r=0.220$, $p<0.001$) between this metric and $\mu$ with a cluster of points in the low-$\mu$ end of the plot, corresponding mostly to uglified images. If we don’t consider this cluster (i.e., unmodified images in picture B) there is no correlation ($r=-0.004$, $p=0.73$). The boxplots (picture C) show that informed observers followed the same trend already present in picture B (unmodified), lowering contrast to uglify and increasing it to beautify the images.

\subsection{Statistical measure of colorfulness (CIELab Gamut Expansion)}

\begin{figure*}[ht]
    \centering
    \includegraphics[width=0.8\textwidth]{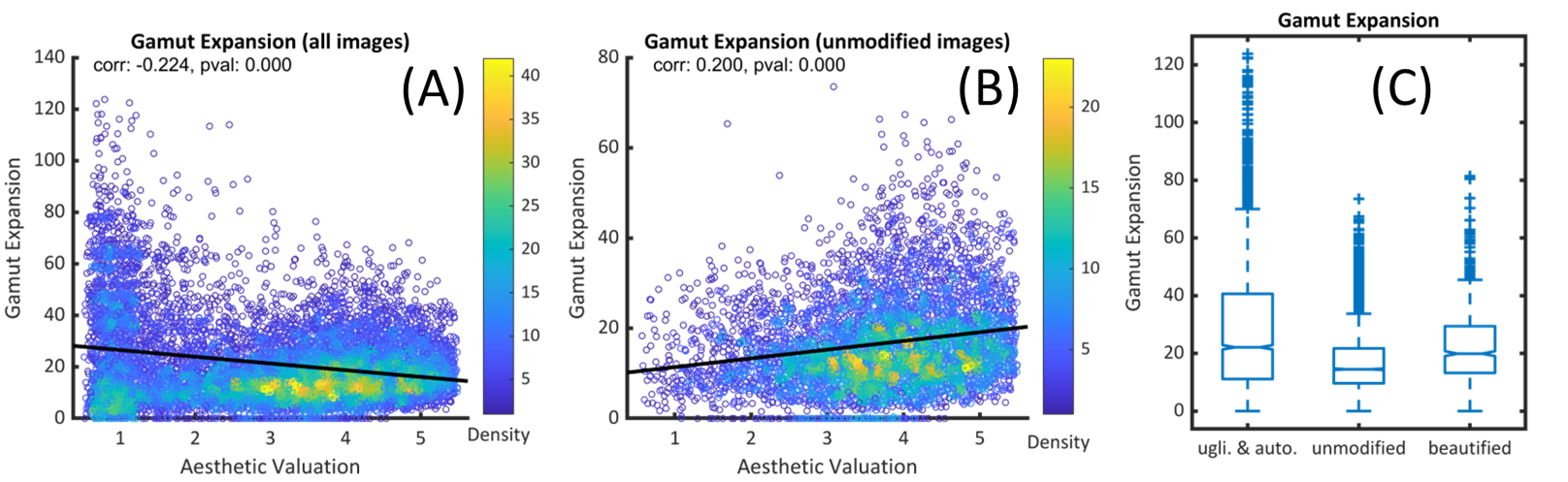}
    \caption{Results for the Gamut Expansion measure of colorfulness. (A) Scatterplot for all images. (B) Same for unmodified images. (C) Metric values for the three groups of images considered. Boxplots as in previous figures.}
    \label{fig:Fig08}
\end{figure*}

Contrary to previous results, the inclusion of modified images changes the shape of the plot and the sign of the correlation (from $r=0.200$ to $r=-0.224$, with $p<0.001$ in both cases). Indeed, Figure~\ref{fig:Fig08} shows a non-monotonous relationship between colorfulness and $\mu$, with a much larger vertical spread of data points for uglified \& auto-uglified images suggesting that informed observers did both, contract and expand the color gamut to lower $\mu$. Figure~\ref{fig:FigA4} in the appendix shows extreme examples of images from the low-$\mu$ end of the plot with both, increased and decreased colorfulness.

\subsection{Overall perceived colorfulness, and saturation from the HSV color model}

Two other metrics related to color, overall colorfulness~\citep{Hasler2003} and saturation from the HSV color model, produced very similar results to those of Figure~\ref{fig:Fig08}. In the case of overall colorfulness, the correlation was negligible ($r=-0.029$, $p<0.001$) for all images, and positive and significant ($r=0.356$, $p<0.001$) for the subset of unmodified images. In the case of saturation, the correlation was negligible ($r=-0.088$, $p<0.001$) for all images and somewhat larger ($r=0.258$, $p<0.001$) for unmodified images. Again, modified, low-$\mu$ images populated the full range of colorfulness on the left side of picture A, in effect making the correlation value negligible. These results are shown in the appendix (Figure~\ref{fig:FigA5} and Figure~\ref{fig:FigA6}).

\subsection{Focus/Blur metric by Thelen \textit{et al}.}

We found no correlation between this metric and $\mu$ ($r=-0.037$, $p<0.001$ and $r=0.025$, $p<0.001$ for all images and unmodified images respectively). A comparison of the data plots shows that, as in the colorfulness case, informed observers freely sharpened or defocused the original images to obtain the corresponding uglified version. These results are shown in the appendix (Figure~\ref{fig:FigA7}).

\subsection{Mean lightness value}

\begin{figure*}[ht]
    \centering
    \includegraphics[width=0.8\textwidth]{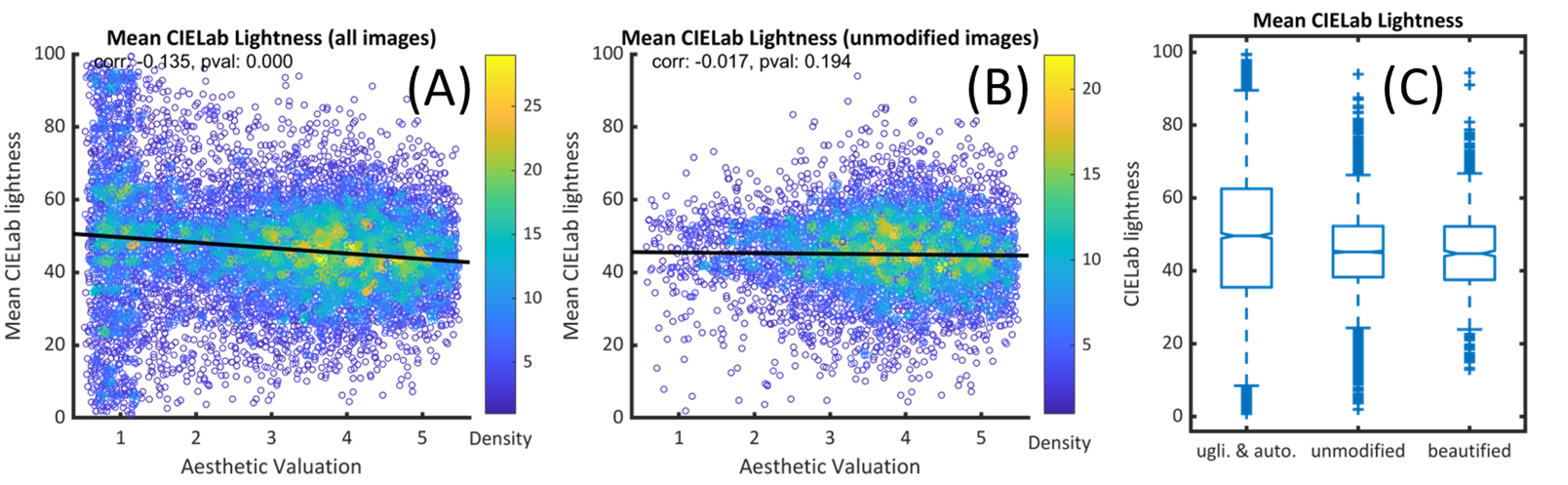}
    \caption{Results for the CIELab “L” value, averaged for each image. (A) Scatterplot for all images. (B) Same for unmodified images. (C) Metric values for the three groups of images considered. Boxplots as in previous figures.}
    \label{fig:Fig09}
\end{figure*}

Both ANOVA and Tukey–Kramer tests confirmed that the means of the boxplots (49.513, 45.001 and 45.495) were significantly different (pairwise $p<0.001$), except for the pair “unmodified”-“beautified” ($p=0.642$). Unmodified images (picture B) show no correlation ($r=-0.017$, $p=0.19$) between these measures. However, informed observers tended to manipulate the whole range of lightness to uglify the images and the overall effect of including modified images was a weak negative correlation ($r=-0.135$, $p<0.001$, picture A). The boxplots in picture C show that, on average, informed observers did not change mean lightness in their attempts to beautify the images.

\subsection{Fourier amplitude slope ($\alpha$)}

We explored the relationship between mean aesthetic valuation ($\mu$) and the Fourier slope ($\alpha$) of our database.

\begin{figure*}[ht]
    \centering
    \includegraphics[width=0.8\textwidth]{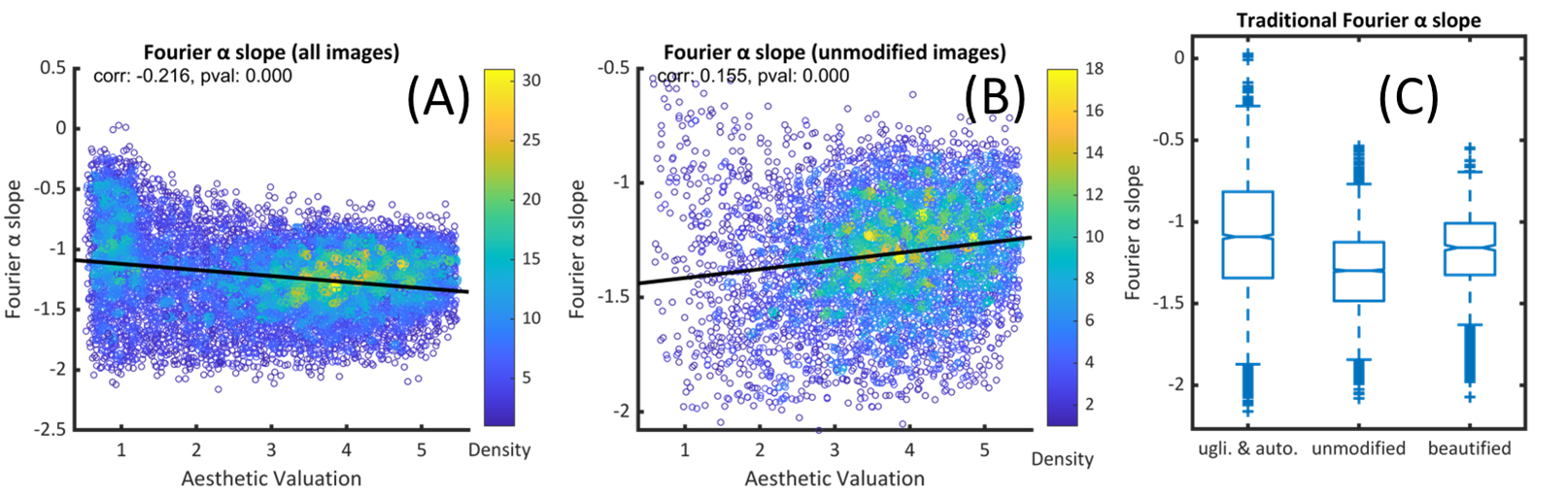}
    \caption{Results for the Fourier amplitude slope ($\alpha$) using the traditional method. (A) Scatterplot for all images. (B) Same for unmodified images. (C) Metric values for the three groups of images considered. Boxplots as in previous figures.}
    \label{fig:Fig10}
\end{figure*}

According to previous studies, most grayscale natural images have slopes in the range $[-1, -1.4]$, and this is also true of our database. Figure~\ref{fig:Fig10} shows that the correlation of $\alpha$ with aesthetic valuation changes sign if both, the uglified and auto-uglified images are considered ($r=-0.216$, $p<0.001$ for all images, and $r=0.155$, $p<0.001$ for unmodified images). Boxplots in picture C show that uglified \& auto-uglified images tended to have a higher $\alpha$ ($\alpha = -1.091$ on average) indicating that informed observers may have increased the Fourier energy content of the images, (e.g., by enhancing the edges) to reduce their aesthetic value.

The results obtained with the unbiased method for calculating $\alpha$ were different. Appendix Figure~\ref{fig:FigA8} shows the same negative sign (unbiased measure) for both correlations ($r=-0.209$, $p<0.001$ for all images and $r=-0.117$ for unmodified images), suggesting a methodology issue.

\subsection{Image complexity metric by Groen \textit{et al}.}

We explored the relationship between mean aesthetic valuation ($\mu$) and image complexity in our database.

\begin{figure*}[ht]
    \centering
    \includegraphics[width=0.8\textwidth]{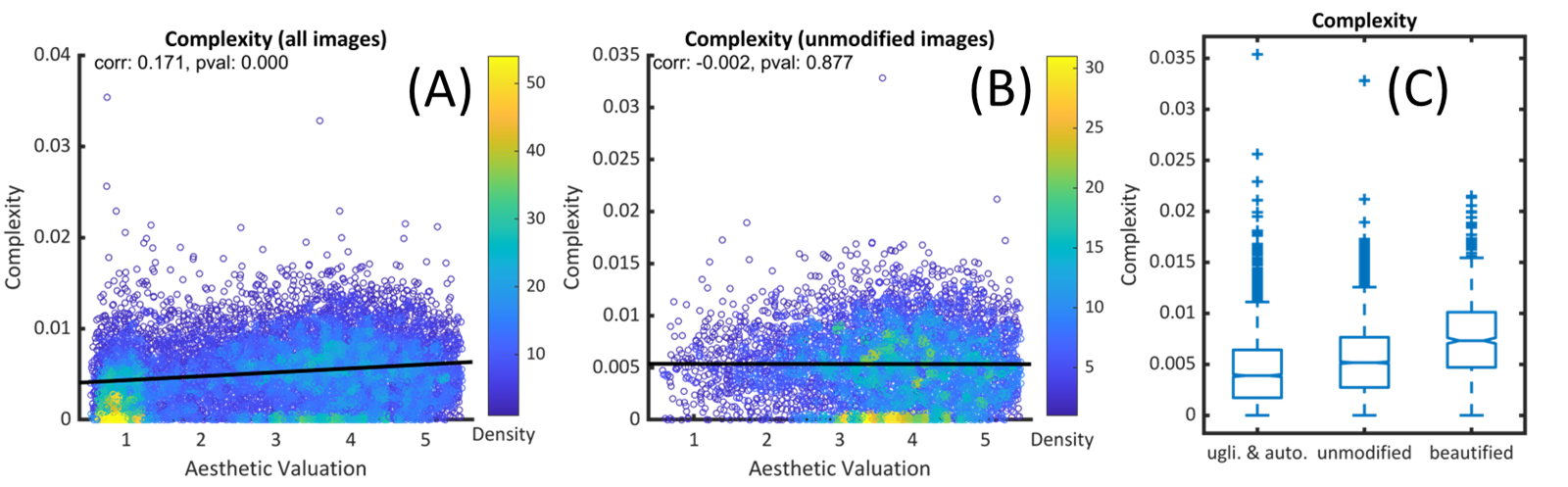}
    \caption{Results for the Image Complexity metric~\citep{Groen2013} considering only Contrast Energy. (A) Scatterplot for all images. (B) Same for unmodified images. (C) Metric values for the three groups of images considered. Boxplots as in previous figures.}
    \label{fig:Fig11}
\end{figure*}

Of the two metrics explored (Contrast Energy and Spatial Coherence), only Contrast Energy (CE) gave meaningful results. Picture A shows a small correlation ($r=0.171$, $p<0.001$) for all images, that is not present for unmodified images in picture B ($r=-0.002$, $p=0.877$). Both ANOVA and Tukey–Kramer tests confirmed that the means of the boxplots (0.004, 0.005 and 0.007) in picture C were significantly different (pairwise $p<0.001$). The significance of these results will be discussed below in the context of broader behavioral performance.

\subsection{Symmetry metric by Brachmann \& Redies}

\begin{figure*}[ht]
    \centering
    \includegraphics[width=0.8\textwidth]{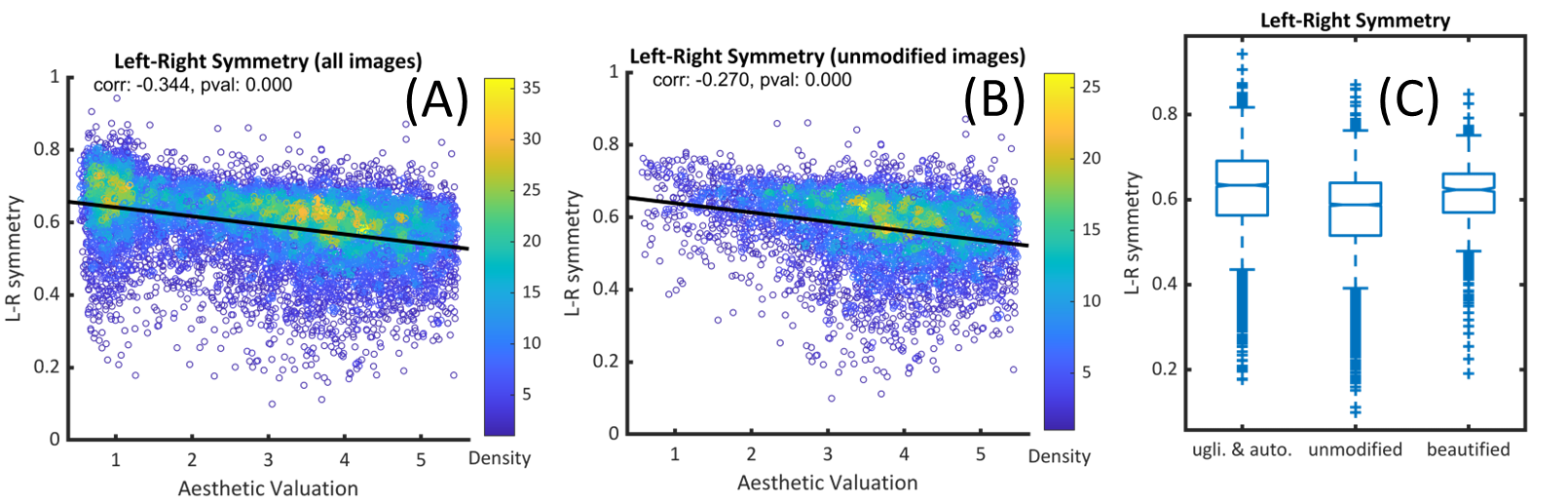}
    \caption{Results for the Left-Right Symmetry metric~\citep{Brachmann2016}. (A) Scatterplot for all images. (B) Same for unmodified images. (C) Metric values for the three groups of images considered. Boxplots as in previous figures.}
    \label{fig:Fig12}
\end{figure*}

Figure~\ref{fig:Fig12} was obtained by plotting the mean of the left-right (L-R) symmetry metric~\citep{Brachmann2016} against $\mu$ for the whole database (picture A), the unmodified images only (picture B), and the boxplots of this metric for three groups (uglified \& auto-uglified, unmodified and beautified in picture C).

Both ANOVA and Tukey–Kramer tests confirmed that the means of the boxplots (0.620, 0.570 and 0.608) were significantly different (pairwise $p<0.001$) except for the pair “ugli \& auto”-“beautified” ($p=0.002$). According to the authors of this metric, larger values correspond to more L-R symmetrical images. We also calculated the same metric for up-down symmetry and the results (shown in Figure~\ref{fig:FigA9} of the appendix) were qualitatively similar. Picture A in Figure~\ref{fig:Fig12} shows an important negative correlation ($r=-0.344$, $p<0.001$) between this metric and aesthetic valuation, indicating that naïve observers gave L-R symmetrical images lower aesthetic valuations. The same is true for unmodified images in picture B ($r=-0.270$, $p<0.001$). Boxplots in picture C also confirmed this tendency, however, it is worth noticing that uglified and beautified images had very close L-R symmetry values on average. We show some extreme examples in the appendix (Fig. A10) and discuss these results in the next section.

\subsection{Mean 3D depth}

We applied the 3D depth algorithm~\citep{Ranftl2021} and calculated its mean value to obtain an estimation of depth in each image. We resized the images so that the longer side was 150 pixels to save computer processing power.

\begin{figure*}[ht]
    \centering
    \includegraphics[width=0.8\textwidth]{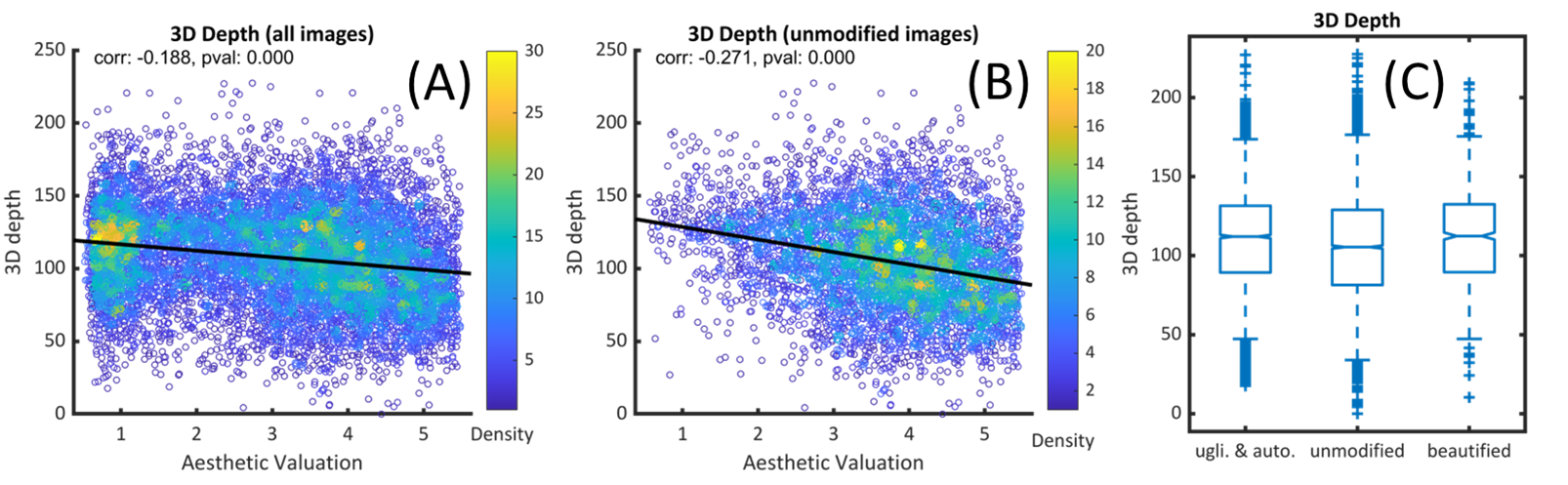}
    \caption{Results for the 3D depth metric (Ranftl \textit{et al}., 2021). (A) Scatterplot for all images. (B) Same for unmodified images. (C) Metric values for the three groups of images considered. Boxplots as in previous figures.}
    \label{fig:Fig13}
\end{figure*}

Figure~\ref{fig:Fig13} shows the results for the mean 3D depth metric as a function of aesthetic valuation for all images (picture A) and unmodified images only (picture B). Picture C shows the boxplots of this metric for three groups (uglified \& auto-uglified, unmodified, and beautified). Both ANOVA and Tukey–Kramer tests confirmed that the means of the boxplots (110.74, 105.38 and 111.94) were significantly different, except for the pair “ugli. \& auto”-“beautified” ($p=0.573$). The results show a small negative correlation ($r=-0.188$, $p<0.001$) between this 3D depth measure and $\mu$ for all images, suggesting that in general, naïve observers prefer open views and landscapes to close-ups. Unmodified images also follow this trend ($r=-0.271$, $p<0.001$). The boxplots in picture C reveal that this metric did not capture the difference between the uglification and the beautification process.

Individuals can access the MSC image dataset, software files, and supplemental files through the OSF Open Science Framework and Share website (\url{https://osf.io/zgsvj/?view_only=8607acce693741c5bb3799b18ed13ca3}).

\section{Discussion}

It may be argued that image aesthetics are intrinsically linked with semantics, so the MSC dataset represents only a subset of the full image spectrum with characteristics that are different from those of everyday images. From our point of view, the MSC dataset can be seen as a controlled experiment designed to isolate perceptual factors from semantic ones, at least for one broad class of images. This reductionist approach is intentional and aligns with methods in neuroscience that separate complex processes to understand individual components, like the "what" (ventral) and the "where" (dorsal) pathways~\citep{Ungerleider1994} or specialized brain regions like visual areas V1, V2, V3~\citep{Felleman1991}, etc. By minimizing semantic factors and holding them constant, we provide a foundation for models that can later introduce semantic complexity. Although our dataset may not capture the full scope of real-world aesthetics, this methodology enables a structured investigation of aesthetic decisions based on perception, paving the way for future research that integrates semantic components and deepens our understanding of image aesthetics.

We started from the supposition that current empirical approaches to visual aesthetics lack a well-balanced, unbiased, and semantics-free database and worked to generate one. The process of creating such a database was incremental and required several attempts before we ended up with a distribution of average aesthetic valuations that contained a balanced number of examples. Several decisions had to be made along the process: (a) the decision to remove human-made objects, people and animals was somewhat arbitrary, but since this was done manually, we felt the need to establish clear and easy-to-follow rules, (b) the decision to use the information contained in valuation histograms to remove the discretization by fitting truncated Gaussians gave us correlations and plots that are easy to analyze visually and (c) the decision to limit the number of original (mostly beautiful) images gave us a more balanced database in terms of its overall content of ugly, intermediate and beautiful examples.

We lack a solid explanation for the low dynamic range and the scarcity of examples categorized as extremely ugly and extremely beautiful examples of the AVA database and other databases derived from the DPChallenge. This may have to do with a combination of psychophysical errors such as the time-order error first described by Fechner~\citep{Stevens1957} and contextual effects like the within-category assimilation effect~\citep{Tajfel1963}. The former refers to the fact that the order of presentation influences how stimuli are judged~\citep{Hellstrom1985}, and the later refers to a decrease in perceived difference that occurs when two stimuli are somehow perceived as belonging to the same category~\citep{Arielli2012}. Since we do not know for sure how the stimuli were presented to DPChallenge participants (e.g., if the images were always presented in the same order or randomized?) and, similarly, do not know the effects of the different categories (“themes” or “challenges”), it is difficult to find a satisfactory answer. Nevertheless, by presenting our images in random order, not having “themes” or “challenges” in our crowdsourcing, and creating our own “ugly” examples, we have avoided at least these two important flaws. In consequence MSC is more balanced, has similar numbers of images in all valuation categories and has no bias towards beautiful images. Indeed, when naïve observers were presented with uglified or auto-uglified images, they tended to rate them with the lowest aesthetic valuations. Figure~\ref{fig:Fig05} shows that modified images are generally considered uglier than unmodified images, filling a void present in classical databases. In other words, when presented with extremely ugly examples, naïve observers rescale their decisions and assign previously low-rated images much higher ratings than otherwise—a consequence of the anchoring effect~\citep{Tversky1974}.

We acknowledge that the distinctions mentioned above (such as semantic-rich images, ratings done under themes, scarcity of extreme examples, and bias towards beauty) make direct comparisons between DPChallenge-derived datasets like AVA and the MSC database at least problematic. Perhaps a more meaningful approach would be to compare MSC with a subset of AVA containing only natural scenes. However, the lack of extreme examples in AVA (there are only 60 images rated between 0 and 1, and only 21 rated between 9 and 10) prevents any insightful comparison, even after removing man-made objects.

Another crucial aspect to consider is the effectiveness of the semantic content removal in MSC. It is difficult to dispute that eliminating elements such as man-made objects, people, animals, etc., would qualitatively result in a dataset that elicits fewer cognitive and/or emotional responses. The challenge here is to assess this affirmation quantitatively, which itself merits a future analysis. Meanwhile, we must acknowledge that we cannot entirely eliminate semantics, and some observers may have a strong attachment, preference, or emotional associations towards natural scenes and objects, as supported by research linking the viewing of natural scenes to improved health/well-being outcomes (for a review, see~\citep{Bratman2019}). Alternatively, we had to define specific criteria to select images lacking semantic content from the extensive range of examples available on the internet. The exclusion of man-made objects, people, etc., despite being arbitrary, proved to be a reasonable approximation. Importantly, the vast majority of manipulated images stayed within the set of natural images in the sense that a naïve observer would classify them as belonging to this set.

We also examined the actions of informed observers tasked with providing ugly and beautiful examples. In particular, the uglification and randomization processes (the latter can be seen as an automatization of the first) gave us further insight about the changes that informed observers implemented to produce ugly images in the absence of strong emotional content. Sometimes these changes follow the trend already present in unmodified images (e.g., with contrast, symmetry, unbiased $\alpha$-slope, and 3D depth), and sometimes they go in the opposite direction (e.g., with colorfulness, saturation, etc.), informing us of a different motivation. Also, it could be argued that the inherent limitations of the "Uglifier" might produce stereotyped results, for instance, consistently generating “ugly” images with unrealistic colors, low saturation, or similar characteristics. To assess this, we compared the low-level features of images rated as "ugly" (aesthetic rating $\mu \leq 2.5$) by our uninformed participants, including both Uglifier-modified and unmodified images. Visual inspection shows that most low-level features fall within a comparable range, with the main exceptions being the color gamut (“gamut expansion” and “colorfulness”). While our informed participants tended to exaggerate color vividness to make images appear “uglier,” this exaggeration was also observed in the color profiles of unmodified images classified as "ugly."

A central finding when comparing the correlations between metrics and aesthetic judgments before and after correcting for the bias towards beautiful images is that the polarity or significance of these correlations may change. This has significant implications for correlational studies in empirical aesthetics that do not account for the typical bias towards one end of the judgment spectrum.

There are, of course, many more possible features than the ones we considered that might give us insight into what influenced decisions during the uglification process and crowdsourcing, but exploring all of them is beyond the scope of our work. Here we want to make a point on the need to separate the actual stimulus from its context and more importantly, to give ugliness and beauty the same relevance when studying visual aesthetics. To this end, we only have to look at Table~\ref{tab:Table05}. For example, the addition of modified images has boosted the correlations with $\mu$ in several cases ($C_{Lab}$, $C_L$ and global contrast; lightness, unbiased $\alpha$-slope, and symmetry). In other cases, it has lessened the correlations (Cab contrast, and 3D depth), or even reversed them (both colorfulness metrics, saturation and traditional $\alpha$-slope). The reduction or even reversal in correlation indicates that informed observers did not follow the general trend (already present in unmodified images) to produce uglified images. Another important feature in Table~\ref{tab:Table05} is that the comparison between unmodified images in the MSC database and the AVA database show that most of these correlations with $\mu$ are not present in a semantic-rich database such as AVA. The only exceptions is $\alpha$-slope (both measures). Each metric has unique characteristics that are worth discussing separately.

\subsection{Contrast}

Artists and image professionals manipulate highlights, shadows, and lighting in relation to their surroundings to create overall effects (gloomy, moody, airy, bright, etc.). The same applies to color, allowing them to create aesthetically pleasing compositions by spatially arranging different colored elements. They exploit a property of the human visual system, which encodes its input by normalizing light intensity and processing mostly contrast patterns. Even though it is unclear how much contrast weighs in the final aesthetic decision, there is consensus on the importance of chromatic and achromatic contrast, and that it needs to be calculated in a multiscale framework for complex images~\citep{Peli1990}. Our results support these points since the highest correlation in Table~\ref{tab:Table05} occurs between multiscale contrast and $\mu$.

\begin{figure*}[ht]
    \centering
    \includegraphics[width=0.8\textwidth]{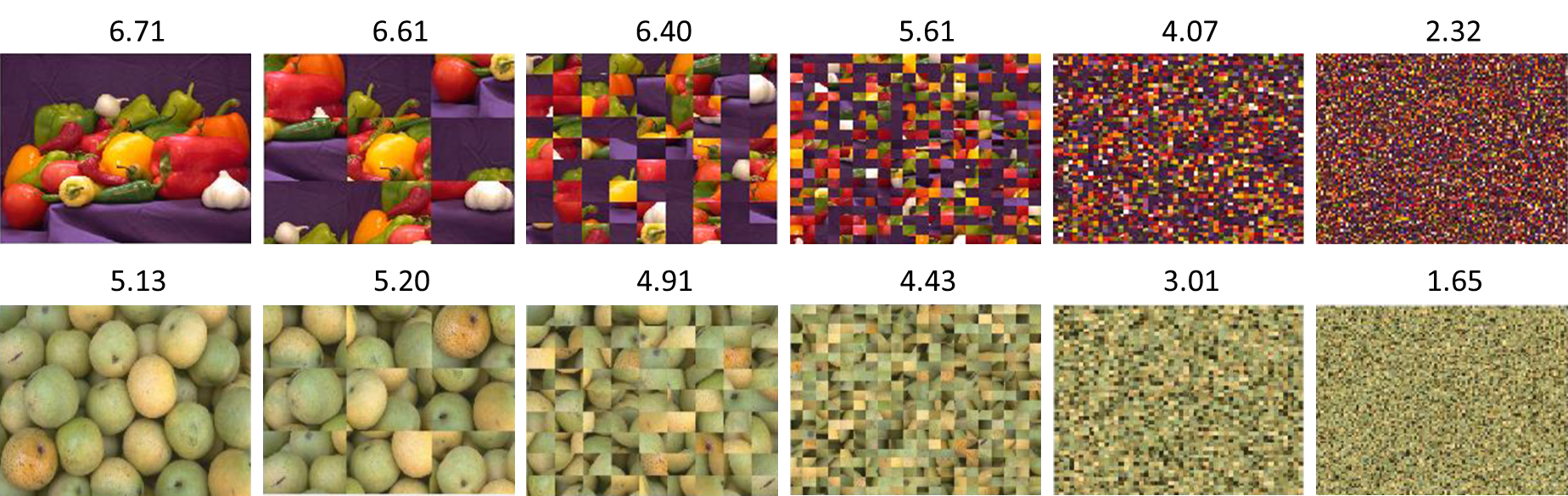}
    \caption{Two examples of how the multiscale contrast value decreases as the image content becomes fragmented. The numbers show the corresponding multiscale contrast value ($C_{Lab}$) for each image.}
    \label{fig:Fig14}
\end{figure*}

Multiscale contrast is also inversely related to image fragmentation (Fig.~14 shows that $C_{Lab}$ decreases almost monotonically as the image gets scrambled). This is a property of the filters used, since the positive and negative parts of large-scale DoGs tend to cancel each other in highly fragmented regions. With that in mind, we can reinterpret Figure~\ref{fig:Fig14} and say that naïve observers showed aesthetic preferences for unfragmented images, which is in line with psychophysical results showing that mean beauty rating decreases with crowding~\citep{Zhou2022}.

Surprisingly, there was no correlation between multiscale contrast and $\mu$ for the AVA database (see Table~\ref{tab:Table05}). We attribute this to the strong semantic content of AVA, as discussed in the methods section, supporting the idea of a reductionist approach to understand the basis of aesthetic predictions. Also, the GCF of Matković \textit{et al}. did not correlate with $\mu$ for our unmodified images nor the AVA database, although informed observers did increase/lower the value of this metric in their attempts to increase/lower $\mu$ (Fig.~7, pictures A and C).

\subsection{Colorfulness}

We expected colorfulness to play a strong role in aesthetic valuations, since vivid colors are commonly used in image enhancement~\citep{Zamir2021, Zamir2017} and colorless pictures of natural objects are usually deemed as “boring” or “uninteresting”. Indeed, that is the case for unmodified images, where all measures of colorfulness (CIELab gamut expansion, HVS saturation and Hasler \& Suesstrunk’s measure) correlate with $\mu$ (see Table~\ref{tab:Table05}). These correlations are absent in the AVA database presumably masked by the semantic content or because DPChallenge users (being mostly amateur photographers) have some kind of bias in this respect. For example, saturated colors are likely to be common in human-made objects and do not have the same impact as in natural objects. The inclusion of modified images reverses the correlation, mostly because informed observers did both, reduce and increase colorfulness to decrease $\mu$. This can be clearly seen for HSV saturation in appendix Figure~\ref{fig:FigA6}, where ugly images fall into two distinct clusters of high and low saturation. Figure~\ref{fig:FigA4} (appendix) shows extreme examples of how informed observers manipulated colorfulness to reduce $\mu$, suggesting perhaps a dislike of extremely unnatural colors of natural objects.

\subsection{Achromatic features}

Although there is an argument to be made of the association of blurred or noisy images with poor technical quality, blurring did not seem to correlate with aesthetic valuation at all. Interestingly, informed observers used both strategies, image blurring and sharpening to lower $\mu$, producing images whose degree of focus deviate significantly from the natural world (see Figure~\ref{fig:FigA7} in the appendix). The same is true for mean CIELab lightness. However, in the case of lightness, informed observers tended to increase rather than decrease the average value (hence the small negative correlation for all images in Table~\ref{tab:Table05}).

\subsection{Fourier amplitude slope ($\alpha$)}

We measured $\alpha$ using two metrics (traditional and unbiased—see methods). The inclusion of modified images produced negative correlations for both metrics ($r=-0.216$, $p<0.001$ for the traditional calculation and $r=-0.209$, $p<0.001$ for the unbiased calculation). We also found that a common strategy for informed observers to uglify (reduce $\mu$) was to increase the value of $\alpha$ (i.e., to make it less negative). In other words, an increase of the Fourier amplitude at high spatial frequencies with the corresponding decrease at low spatial frequencies made images ugly. The results for the unmodified images were not clear. The traditional approach to calculating $\alpha$ resulted in a small positive correlation ($r=0.155$, $p<0.001$ see Figure~\ref{fig:Fig10}) and the unbiased approach resulted in a small negative correlation ($r = -0.117$, $p<0.001$ see appendix Figure~\ref{fig:FigA8}), more in tune with the behavior of the informed observers who manipulated the images. This hints towards a methodological issue: the traditional metric might be overemphasizing Fourier content in the high-spatial frequency side of the spectrum. Interestingly, $\alpha$-slope is one of the few metrics that correlates with $\mu$ in the AVA database ($r=0.227$, $p<0.001$ for the traditional and $r=0.174$, $p<0.001$ for the unbiased). This preference for “sharpened” images over “blurred” images may have to do with the idiosyncrasies of the AVA dataset and the likings of DPChallenge participants.

We also calculated the sum of the residuals obtained from fitting a line with 1/f slope in log-log space to each image. This value ought to be zero for images whose Fourier amplitude profile is exactly equal to a line with slope -1 in log-log Fourier space and increase for less “naturalistic” images. Following this we calculated how this value varied as a function of $\mu$ for all images and for unmodified images in the MSC database. Our results show very small correlations ($r=0.088$, $p<0.001$ and $r=0.045$, $p=0.001$ respectively) between $\mu$ and the sum of the residuals.

\subsection{Image complexity metric by Groen \textit{et al}.}

Given that both metrics, CE and SC, extract contrast at various scales and pool their results, we expected their correlations with aesthetic valuation (see Figure~\ref{fig:Fig11}) to be as high as our parsimonious multiscale contrast metric. These results might indicate that the neural mechanisms involved in the two tasks (aesthetic valuation and man-made versus natural object discrimination) are dependent on different image features, which are differently exploited by the algorithms (as the mild correlation between these two metrics in Figure~\ref{fig:FigA12} indicates).

\subsection{Machine-learnt features}

Fig.~12 shows a clear tendency for naïve observers to give lower $\mu$ values to more symmetrical images. This is true for both L-R symmetry and U-D symmetry (see appendix Figure~\ref{fig:FigA9}) and the literature in this respect is controversial. On the one hand, researchers have repeatedly shown symmetry to play a central role in preference and beauty judgments of visual stimuli~\citep{al-Rifaie2017}, in particular of human faces~\citep{Rhodes2006, Tinio2013}, as this preference has been associated with advantages regarding mate choice~\citep{Thornhill1997}. Symmetry is also a strong determinant of aesthetic judgments of meaningless, abstract patterns and is often explained with reference to the efficient functioning of the human visual system~\citep{Reber2004}. On the other hand, significant differences in the valuation of symmetry has been reported between groups of art experts and non-experts~\citep{Leder2019}. A careful examination of the examples in appendix Figure~\ref{fig:FigA10}, shows that these results may have been a consequence of how the neural net was trained. Images at the top contain objects to one side of the scene, which in the context of music CD covers were considered highly L-R asymmetrical. Images at the bottom have low contrast (most of their features were removed), which makes them highly symmetrical and, according to our results, ugly (low $\mu$). Also, images of extensive foliage with uninteresting features may have been classified as symmetrical and low $\mu$. A cross-correlation analysis (see below) reveals that this metric is significantly correlated to others: multiscale contrast or image fragmentation ($r=-0.39$, $p<0.001$), Focus/Blur ($r=0.44$, $p<0.001$) and $\alpha$-slope ($r=0.36$, $p<0.001$), which helps to understand these results.

The last of the machine-learnt features analyzed was the mean 3D depth metric of Ranftl \textit{et al}.~\citep{Ranftl2021}. Figure~\ref{fig:FigA11} in the appendix shows a few examples of this metric, where high values correspond to content located close to the viewer (and on the same focal plane), and low values correspond to far away objects, like clouds. Our results show that informed observers did follow the general trend in unmodified images by increasing this value to lower $\mu$. Again, there is no correlation between this metric and $\mu$ for the AVA database.

\subsection{Correlations among the different metrics}

Throughout this work we reported Pearson’s $r$ correlation coefficient of quantities such as contrast, colorfulness, focus, $\alpha$-slope, etc., with $\mu$. However, there are probably important correlations between the different metrics in our database. Figure~\ref{fig:FigA12} shows the matrix of correlation coefficients for the results presented above (all images). Their corresponding $p$-values are shown in the appendix (Fig.~A12). The figure has been color-coded to highlight positive (yellowish-green) or negative (orange-red) correlations values. Results with less than 95\% significance level are shown within a black box. All pairwise results are for the MSC database, except the bottom rows which correspond to the correlations already reported in Table~\ref{tab:Table05} between the different metrics and $\mu$ for both the MSC and the AVA databases.

There are some expected correlations in Figure~\ref{fig:FigA12}. For example, metrics of contrast and colorfulness are highly correlated among themselves. Equally, we expected Fourier $\alpha$-slope to be correlated to the Focus/Blur metric of Pertuz \textit{et al}. Multiscale contrast ($C_{Lab}$, which is also linked to image fragmentation, see Figure~\ref{fig:Fig14}) is strongly correlated to the colorfulness metric of Hasler \& Suesstrunk and the L-R symmetry metric of Brachmann \& Redies. This may have to do with the convolutional nature of the neural networks and the databases used in their learning algorithms. It is more surprising to find a correlation between Focus/Blur and Fourier $\alpha$-slope with the L-R Symmetry metric of Brachmann and Redies. The $p$-values for all significant correlations were very small.

The removal of semantic-rich images is likely to reduce the generalizability of the MSC database, limiting it to study the contribution of the sensory–motor (perceptual) circuitry~\citep{Chatterjee2014, Redies2015} to aesthetic judgment. Nonetheless, we believe that using a reductionist approach, such as the one proposed, is a good and essential step towards advancing understanding when facing notably intricate and complex problems. In the future we may want to explore the contribution of the meaning–knowledge and the emotion–valuation circuitry~\citep{Chatterjee2014, Redies2015} for generating ugly imagery using a similar approach. Many questions remain. For example, what is the role of emotion in the uglification process? Will informed observers implement changes similar to those we uncovered in the presence of strong, emotion-driving, semantic content?

% \begin{figure*}[ht!]
%     \centering
%     \includegraphics[width=0.8\textwidth]{Fig.15_Pearsonr_Correlations.png}
%     \caption{Matrix of Pearson’s $r$ correlation coefficients for the different metrics presented above. The picture shows correlations for the pair of metrics determined by the rows and columns. The matrix has been color-coded to indicate positive or negative correlation (green-red) with values below the 95\% significance level inside a black box. The bottom rows correspond to the correlations already reported in Table~\ref{tab:Table05} between the different metrics and $\mu$ for both the MSC and the AVA databases. Appendix Figure~A12 shows the same matrix with its corresponding numerical $p$-values.}
%     \label{fig:Fig15}
% \end{figure*}

\section{Conclusions}

We created a database of aesthetically evaluated images to allow us to investigate the contribution of image characteristics to aesthetic valuations within the domain of computational aesthetics. Our motivation was to address three biases in existing databases (consisting of publicly rated images derived from online platforms), namely (a) a general bias towards beauty, i.e., ugly images tend not to be included, (b) a general bias towards mid-range valuations, i.e., very few images rated as “extremely beautiful” or “extremely ugly,” and (c) their valuation made within voting contests under certain “themes,” i.e., influenced by the semantic categories applied to images, hence featuring a semantic context/bias. To minimize the bias towards beauty and the lack of extreme examples we let participants systematically manipulate spatiochromatic characteristics of images to either create beautiful or ugly versions. To minimize the effect of the semantic bias we restricted the content of our database to objects from the natural environment. The manipulations were recorded and then randomly applied to a subset of images. All three types of manipulated images were added to a set of unmodified images resulting in a more balanced dataset of over 10,000 images. A subsequent comparison of the final database, a subset including only unmodified images, and a commonly used database, showed differences in selected image-metrics and their relation to aesthetic valuations.

Our findings show that collecting images solely from the internet can result in databases with limited aesthetic range, that are strongly biased towards beauty. This led us to the central finding of this study: when comparing the correlations between metrics and aesthetic judgments before and after correcting for the bias towards beautiful images, the polarity or significance of these correlations may change. This has significant implications—and serves as a cautionary note—for correlational studies in empirical aesthetics that fail to account for the typical bias towards one end of the judgment spectrum. Conducting correlative analyses on such databases may indeed result in spurious associations between image features and aesthetic judgements. By contrast, enlarging the range of aesthetic valuations can modify, and sometimes revert, associations between image metrics and observers’ responses. Additionally, removing semantic content unveils the presence of many correlations, which are generally absent in semantic-rich databases. These results have broad implications for our understanding of the relationships between stimuli features and aesthetic valuations in general, as well as for interpreting the existing literature on the topic.

%\clearpage % To flush out all pending floats before the new page.

\onecolumn
\section*{Acknowledgments}

Supported by the Ministerio de Ciencia e Innovación, Gobierno de España MCIN/AEI/10.13039/501100011033: grants PID2020-118254RB-I00 and TED2021-132513B-I00, by the Agencia de Gestió d’Ajuts Univesitaris i de Recerca (AGAUR) through 2021-SGR-01470, and CERCA Programme / Generalitat de Catalunya.

Supported by PID2019-104174GB-I00, funded by MCIN/AEI/10.13039/501100011033, Spain.

Funded by a Maria Zambrano Fellowship for the attraction of international talent for the requalification of the Spanish university system—NextGeneration EU (ALRC).

%\clearpage % To flush out all pending floats before the new page.
\newpage

\appendix
\renewcommand{\thefigure}{A\arabic{figure}}
\section*{Appendix A}
\setcounter{figure}{0}  % Reset figure counter

\subsection*{A.0. Equations used to obtain the multiscale contrast}
\label{eq:contrast}

Chromatic and achromatic multiscale contrast was calculated as follows:

\begin{align}
C_{\text{Lab}} &= \frac{1}{N_\lambda} \cdot \frac{1}{xy} \sum_\lambda \sum_{x,y} \sqrt{
    \left(D_\lambda(x,y) \cdot L(x,y)\right)^2 +
    \left(D_\lambda(x,y) \cdot a(x,y)\right)^2 +
    \left(D_\lambda(x,y) \cdot b(x,y)\right)^2
} \label{eq:contrast1} \\
C_{ab} &= \frac{1}{N_\lambda} \cdot \frac{1}{xy} \sum_\lambda \sum_{x,y} \sqrt{
    \left(D_\lambda(x,y) \cdot a(x,y)\right)^2 +
    \left(D_\lambda(x,y) \cdot b(x,y)\right)^2
} \label{eq:contrast2} \\
C_L &= \frac{1}{N_\lambda} \cdot \frac{1}{xy} \sum_\lambda \sum_{x,y} \sqrt{
    \left(D_\lambda(x,y) \cdot L(x,y)\right)^2
} \label{eq:contrast3} 
\end{align}

where $x, y$ represents the pixel coordinates, $L, a$ and $b$ represent the three CIELab planes, $D$ represents the difference-of-Gaussians (DoG) operator, $l$ represents a particular spatial scale, $N_\lambda$ is the total number of spatial scales for the image and $\cdot$ represents a mathematical convolution.

\subsection*{A.1. Alternative methods for computing the final aesthetic valuations ($\mu$)}

\begin{figure}[ht]
  \centering
  \includegraphics[width=0.8\textwidth]{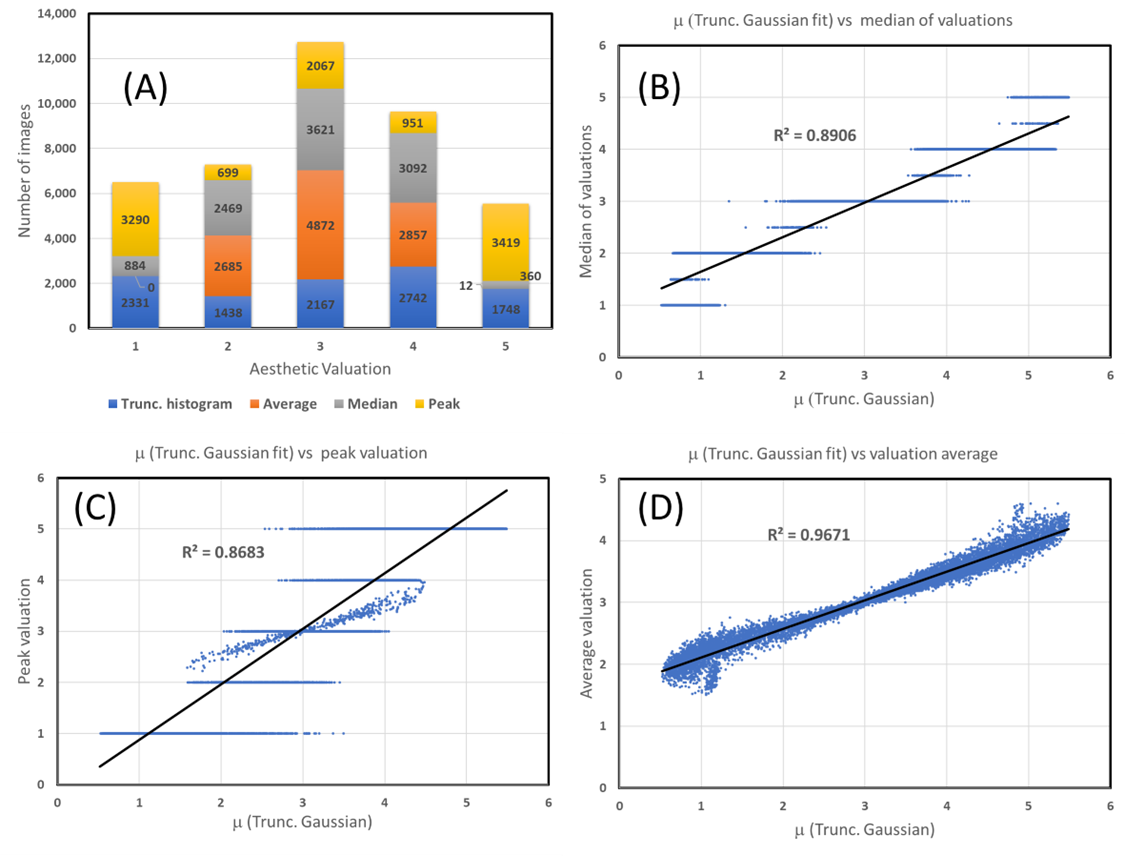}
  \caption{Comparisons between truncated gaussian ($\mu$) and three alternative methods for computing the final aesthetic valuations (valuation average, valuation median and peak valuation). (A) Composite histogram showing the number of images in each of the five Likert scale categories obtained by each of the four methods (blue= truncated gaussians, orange= average, grey= median and yellow= peak). (B) Relationship between the statistical median and $\mu$. (C) and (D) show the same for peak valuation and average valuation respectively. The non-discrete points of picture (C) reflect the fact that, when the maximum of observer valuations (peak value) corresponded to more than one Likert scale option, we adopted the average. The corresponding coefficient of determination (R$\textsuperscript{2}$) is indicated in each of the plots.}
  \label{fig:FigA1}
\end{figure}
\clearpage % To flush out all pending floats before the new page.

\subsection*{A.2. Purely achromatic multiscale contrast ($C_L$)}

\begin{figure}[hp]
    \centering
    \includegraphics[width=0.75\textwidth]{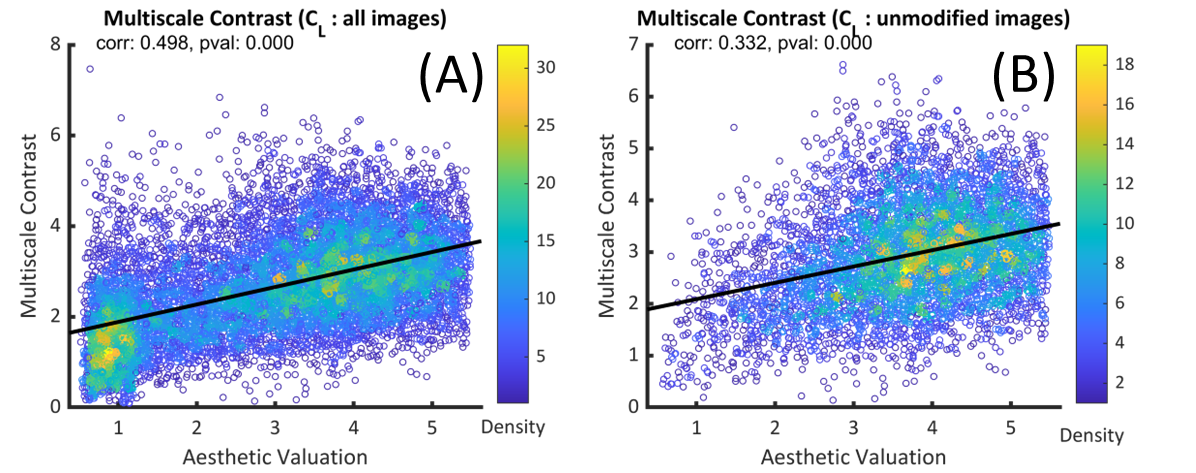}
    \caption{Results from the multiscale contrast ($C_L$), considering only CIELab achromatic channel “L”. Left: scatterplot for all images. The color shows the density of points, with a yellower color corresponding to a higher density and a bluer color to a lower density. Right: unmodified images (i.e., excluding uglified, beautified and auto-uglified). Color code as in the first picture.}
    \label{fig:FigA2}
\end{figure}
%\FloatBarrier

\subsection*{A.3. Purely chromatic multiscale contrast ($C_{ab}$)}

\begin{figure}[H]
    \centering
    \includegraphics[width=0.75\textwidth]{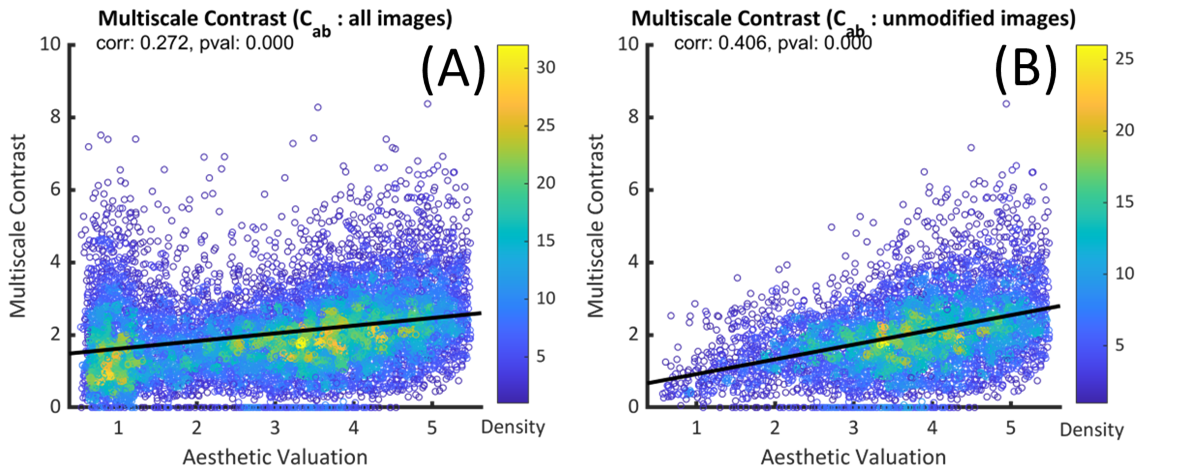}
    \caption{Results from the multiscale contrast ($C_{ab}$), considering both CIELab chromatic channels “a” and “b”. (A) Scatterplot for all images. (B) Same for unmodified images.}
    \label{fig:FigA3}
\end{figure}
\clearpage % To flush out all pending floats before the new page.

\subsection*{A.4. Extreme examples of uglified images}

\begin{figure}[H]
    \centering
    \includegraphics[width=1\textwidth]{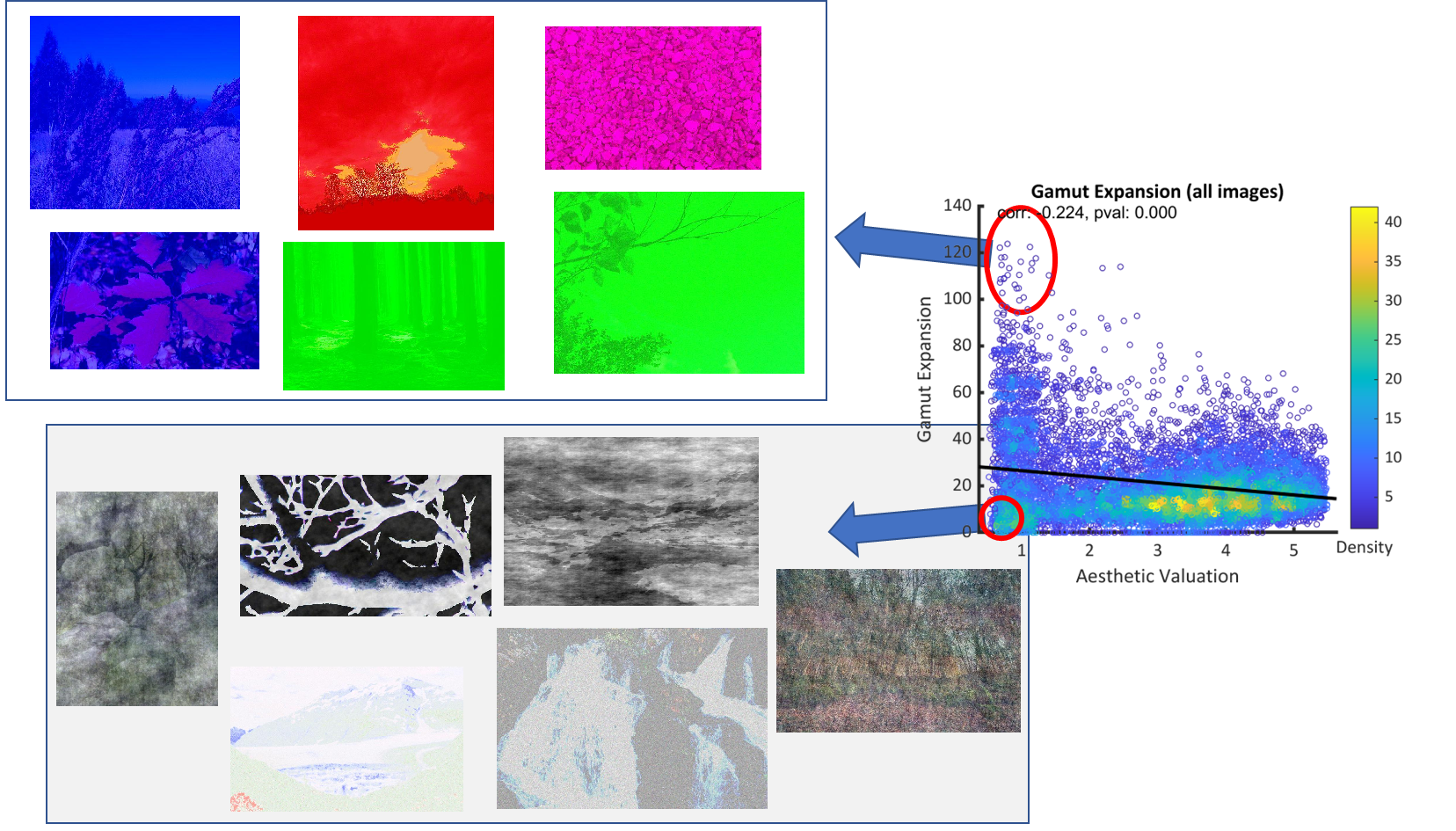}
    \caption{Twelve extreme examples of low aesthetic value images that have either a large (top panel) or a small (bottom panel) CIELab gamut. All images that had their colors saturated or desaturated by informed observers and measured according to the Gamut Expansion metric. Red circles on the scatterplot show the approximate color regions where these images belong.}
    \label{fig:FigA4}
\end{figure}

\subsection*{A.5. Overall Perceived Colorfulness by Hasler \& Suesstrunk}

\begin{figure}[H]
    \centering
    \includegraphics[width=0.75\textwidth]{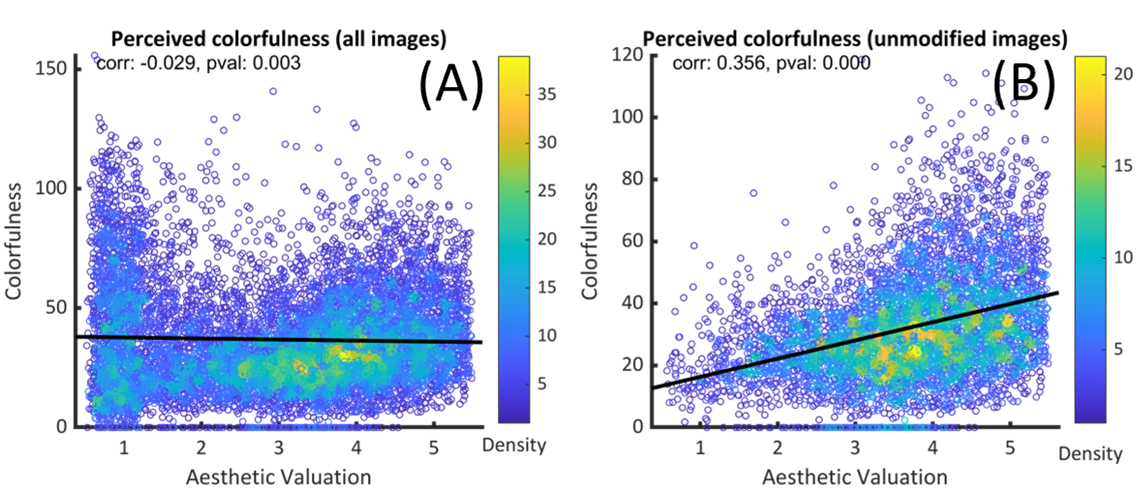}
    \caption{Results for the Overall Colorfulness metric~\citep{Hasler2003}. (A) Scatterplot for all images. (B) Same for unmodified images.}
    \label{fig:FigA5}
\end{figure}

\clearpage % To flush out all pending floats before the new page.

\subsection*{A.6. Saturation from the HSV color model}

\begin{figure}[H]
    \centering
    \includegraphics[width=0.75\textwidth]{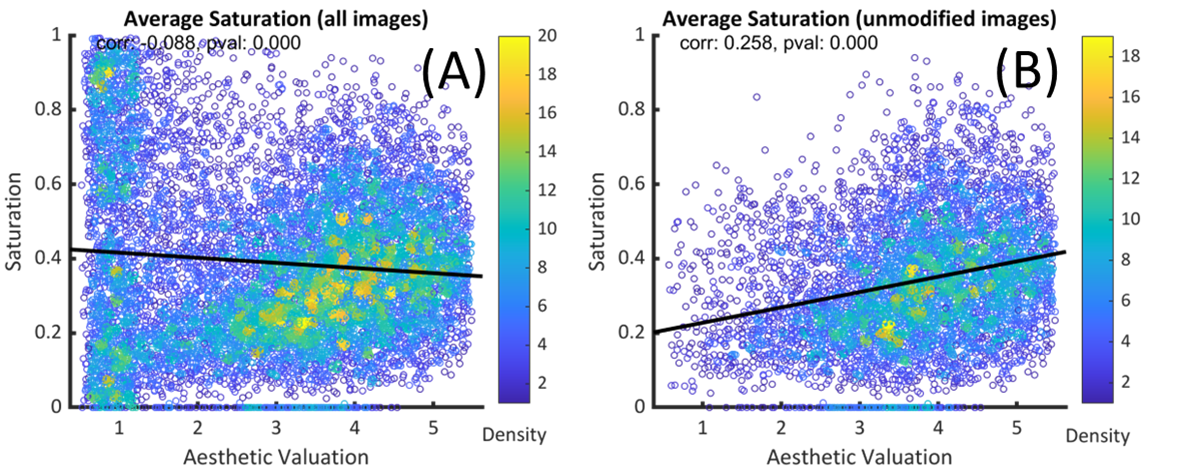}
    \caption{Results for HSV Saturation. (A) Scatterplot for all images. (B) Same for unmodified images.}
    \label{fig:FigA6}
\end{figure}

\subsection*{A.7. Focus/Blur metric by Thelen \textit{et al}.}

\begin{figure}[H]
    \centering
    \includegraphics[width=0.75\textwidth]{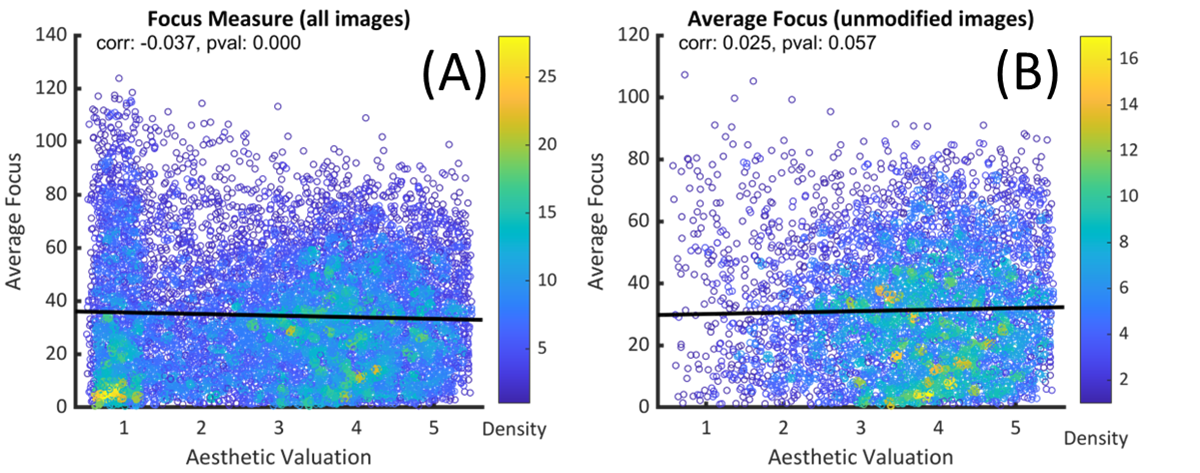}
    \caption{Results for the Focus/Blur metric~\citep{Pertuz2022}. (A) Scatterplot for all images. (B) Same for unmodified images.}
    \label{fig:FigA7}
\end{figure}

%\clearpage % To flush out all pending floats before the new page.

\subsection*{A.8. Fourier amplitude slope ($\alpha$) using the unbiased method}

\begin{figure}[H]
    \centering
    \includegraphics[width=0.75\textwidth]{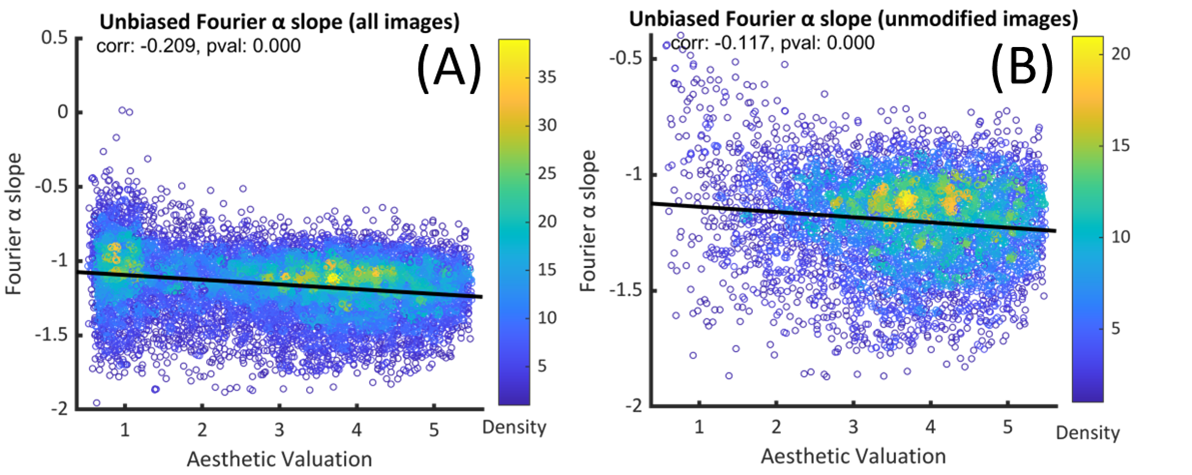}
    \caption{Results for the Fourier amplitude slope ($\alpha$) calculated with the unbiased method. (A) Scatterplot for all images. (B) Same for unmodified images.}
    \label{fig:FigA8}
\end{figure}

\subsection*{A.9. Up-Down Symmetry by Brachmann \& Redies}

\begin{figure}[H]
    \centering
    \includegraphics[width=0.75\textwidth]{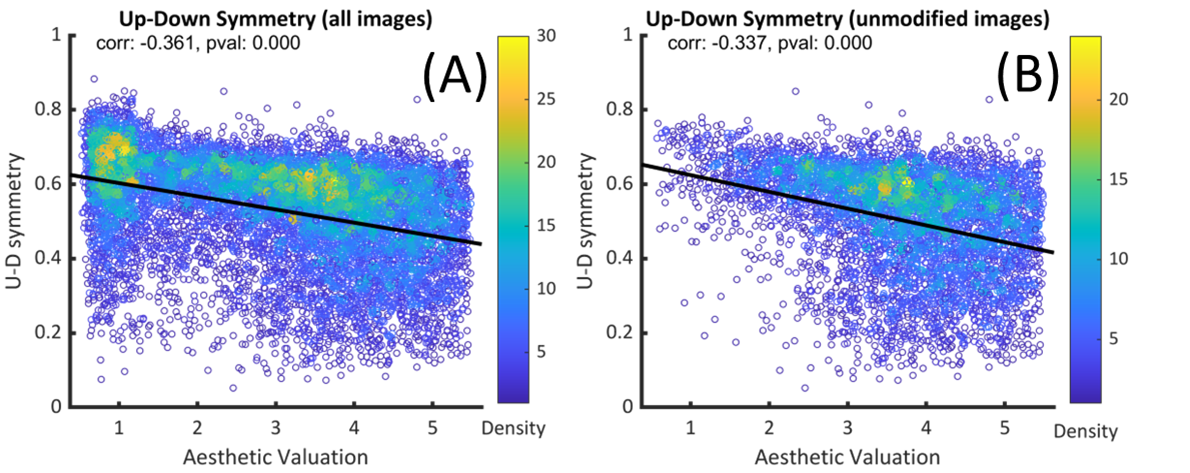}
    \caption{Results for the Up-Down Symmetry metric~\citep{Brachmann2016}. (A) Scatterplot for all images. (B) Same for unmodified images.}
    \label{fig:FigA9}
\end{figure}

%\clearpage % To flush out all pending floats before the new page.

\subsection*{A.10. Extreme examples of poor and good L-R symmetry}

\begin{figure}[H]
    \centering
    \includegraphics[width=1\textwidth]{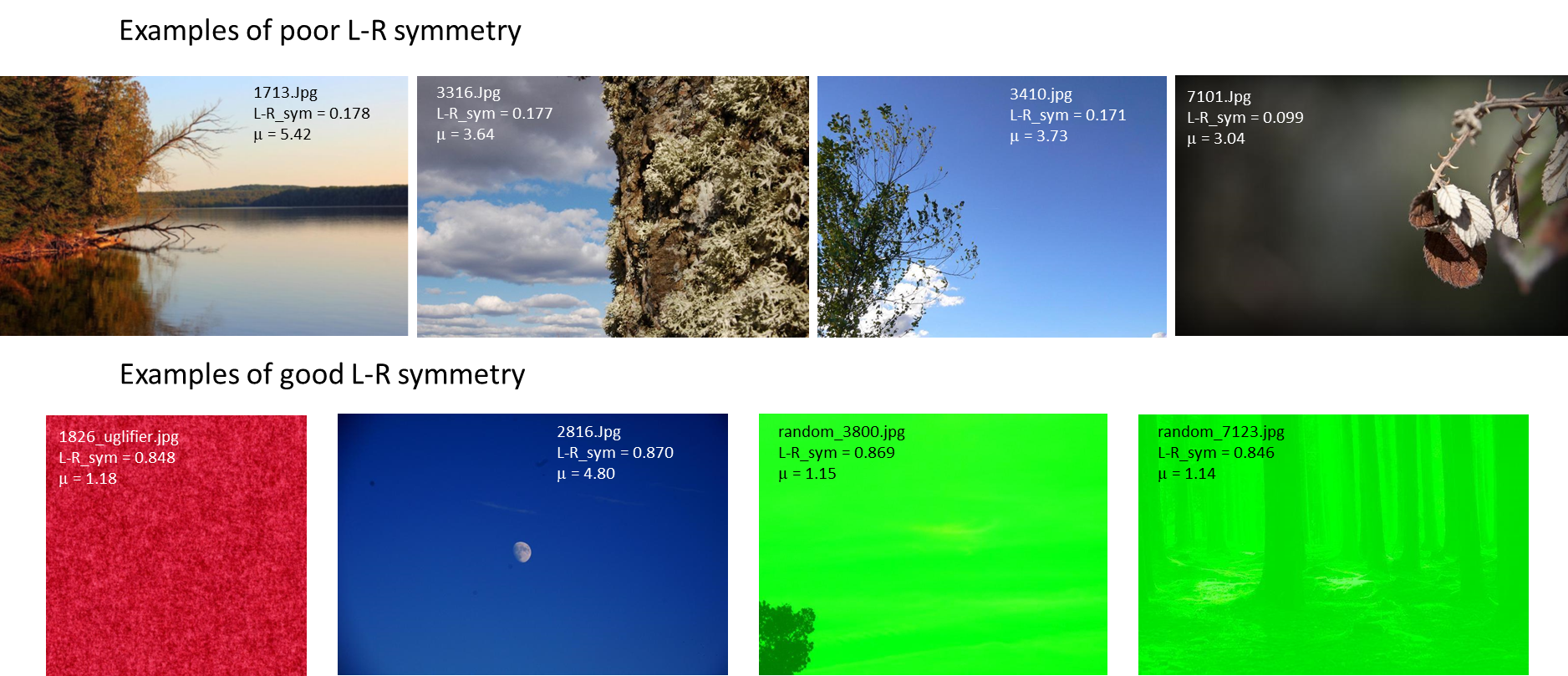}
    \caption{Some extreme examples of L-R symmetry and their aesthetic valuations ($\mu$). The top row shows images labelled as asymmetric (low L-R symmetry) and the bottom row shows images labelled as symmetric (high L-R symmetry) by the algorithm. All images include their results in terms of the L-R symmetry metric and their aesthetic valuations ($\mu$) by the observers.}
    \label{fig:FigA10}
\end{figure}

\clearpage % To flush out all pending floats before the new page.

\subsection*{A.11. Extreme examples of mean 3D depth according to the Ranftl \textit{et al}. metric}

\begin{figure}[H]
    \centering
    \includegraphics[width=1\textwidth]{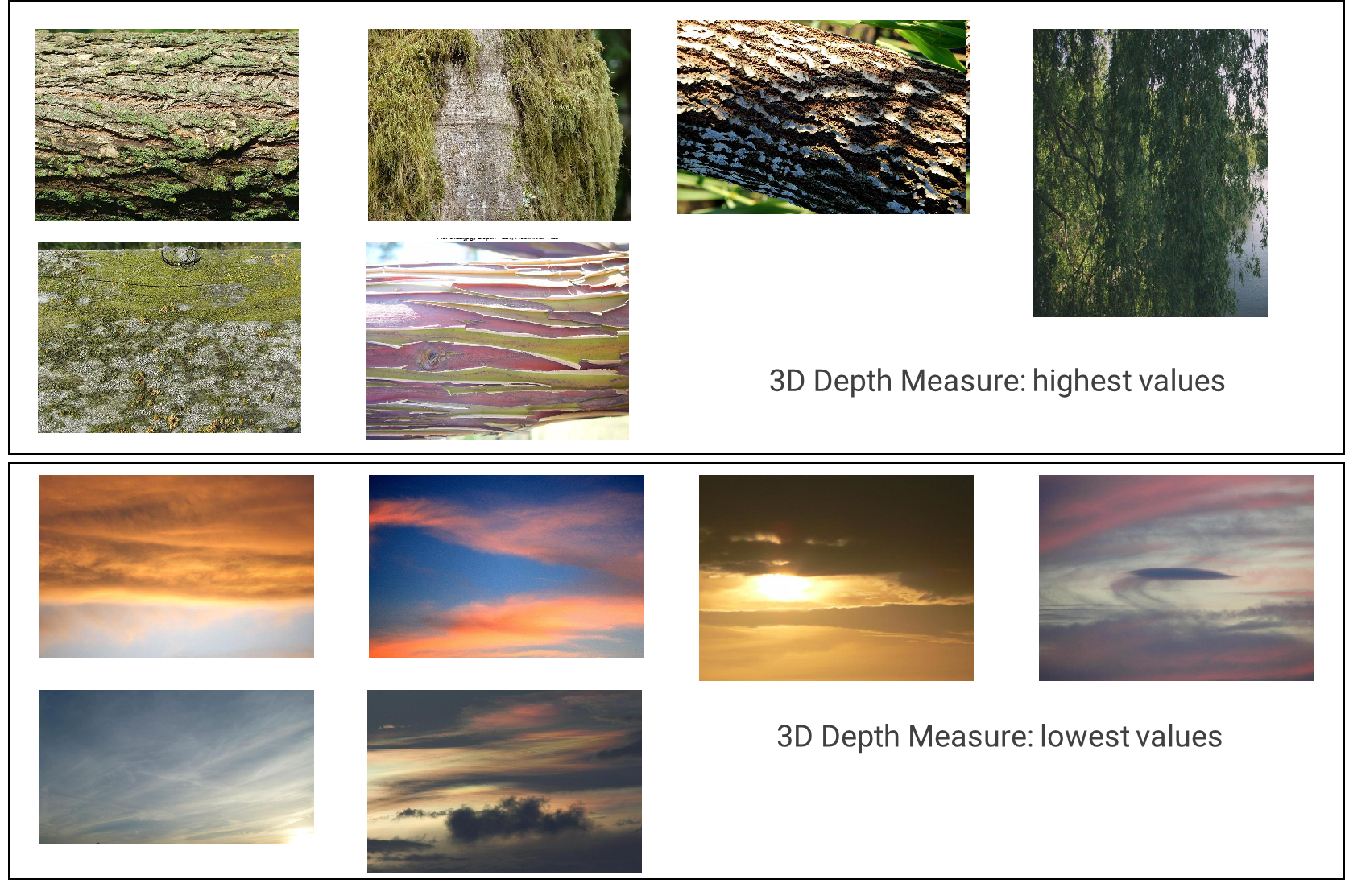}
    \caption{Examples of images with very high and very low 3D depth values according to Ranftl \textit{et al}~\citep{Ranftl2021}. Top panel: images with high 3D depth values. Bottom panel: images with low 3D depth values. Note: in this metric, small values indicate larger mean depth.}
    \label{fig:FigA11}
\end{figure}

\clearpage % To flush out all pending floats before the new page.

\subsection*{A.12. Correlations among the different metrics and p-values}

\begin{figure}[H]
    \centering
    \includegraphics[width=0.75\textwidth]{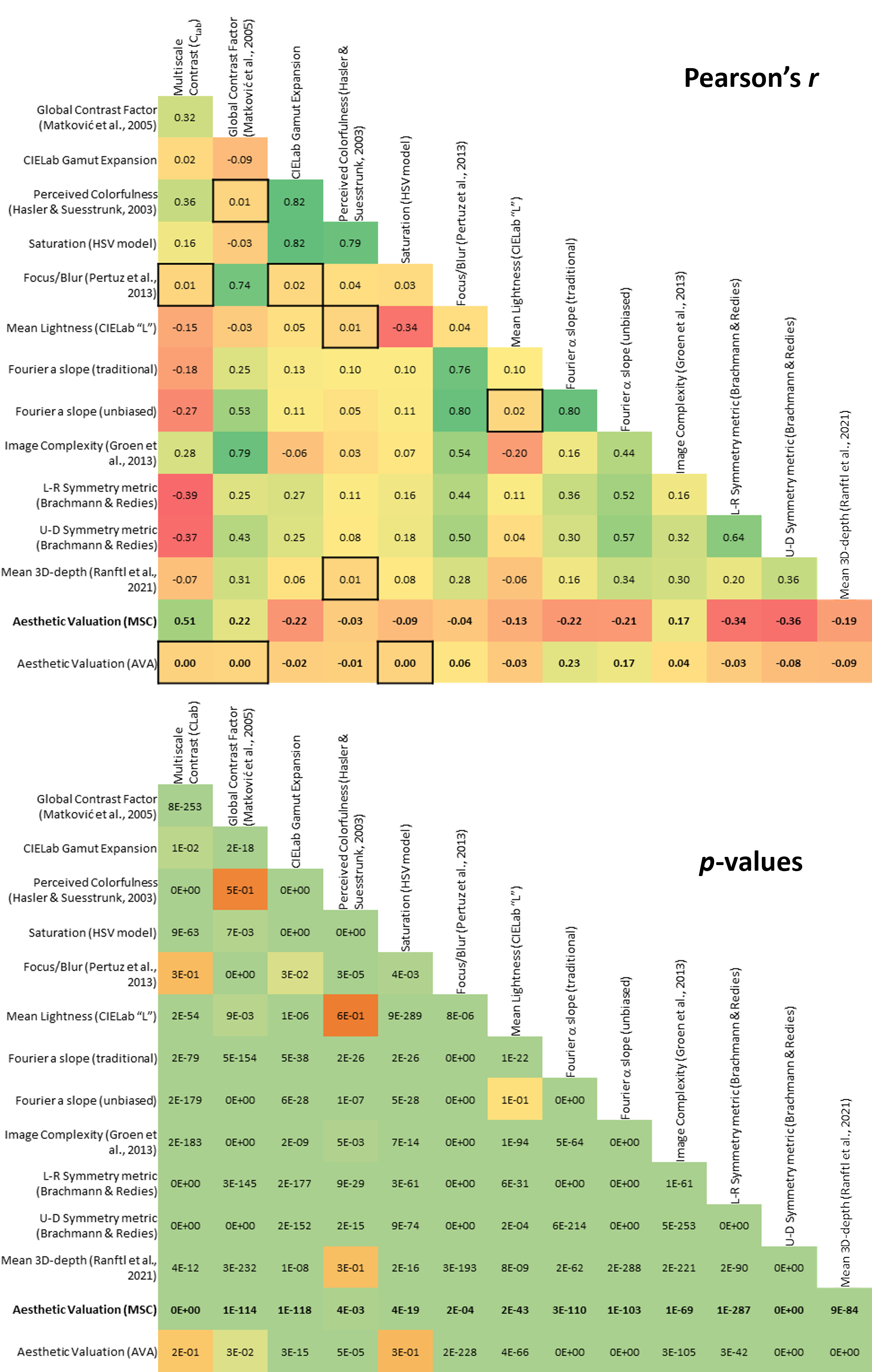}
    \caption{Matrix of Pearson’s $r$ correlation coefficients for the different metrics studied. The top picture shows correlations for the pair of metrics determined by the rows and columns. The matrix has been color-coded to indicate positive or negative correlation (green-red) with values below the 95\% significance level inside a black box. The bottom picture shows the corresponding $p$-values. The matrix has been color-coded to indicate $p$-value size (green for values above the 95\% significance level and yellow/orange for values below it).}
    \label{fig:FigA12}
\end{figure}

\twocolumn

%Bibliography

\bibliographystyle{unsrt}  
\bibliography{references}   % without the .bib extension

\end{document}